\documentclass[a4paper,fleqn]{cas-sc}

\usepackage{amssymb}% http://ctan.org/pkg/amssymb
\usepackage{pifont}% http://ctan.org/pkg/pifont
\newcommand{\cmark}{\ding{51}}%
\newcommand{\xmark}{\ding{55}}%
\usepackage[numbers]{natbib}
\def\tsc#1{\csdef{#1}{\textsc{\lowercase{#1}}\xspace}}
\tsc{WGM}
\tsc{QE}
\begin{document}
\let\WriteBookmarks\relax
\def\floatpagepagefraction{1}
\def\textpagefraction{.001}

\shorttitle{DeepCover: RNN Test Coverage and Error Prediction}    

\shortauthors{Golshanrad and Faghih}

\title [mode = title]{DeepCover: Advancing RNN Test Coverage and Online Error Prediction using State Machine Extraction}  

\tnotemark[1]

% \tnotetext[1]{This work was supported by the University of Tehran.}

\let\printorcid\relax % Remove ORCID footnote

\author[inst1]{Pouria Golshanrad}

\cormark[1]

\fnmark[1]

\ead{pouria.golshanrad@ut.ac.ir}

\ead[url]{https://drts.ut.ac.ir/?page_id=481}

% \credit{<Credit authorship details>}

\affiliation[inst1]{organization={University of Tehran},
            city={Tehran},
            country={Iran}}

\author[inst1]{Fathiyeh Faghih}

\cormark[1]

% Footnote of the second author
\fnmark[2]

% Email id of the second author
\ead{f.faghih@ut.ac.ir}

\ead[url]{https://ece.ut.ac.ir/en/~f.faghih}


\cortext[cor1]{Corresponding author}


\begin{abstract}
Recurrent neural networks (RNNs) have emerged as powerful tools for processing sequential data in various fields, including natural language processing and speech recognition. However, the lack of explainability in RNN models has limited their interpretability, posing challenges in understanding their internal workings. To address this issue, this paper proposes a methodology for extracting a state machine (SM) from an RNN-based model to provide insights into its internal function. The proposed SM extraction algorithm was assessed using four newly proposed metrics: Purity, Richness, Goodness, and Scale. The proposed methodology along with its assessment metrics contribute to increasing explainability in RNN models by providing a clear representation of their internal decision making process through the extracted SM. In addition to improving the explainability of RNNs, the extracted SM can be used to advance testing and and monitoring of the primary RNN-based model. To enhance RNN testing, we introduce six model coverage criteria based on the extracted SM, serving as metrics for evaluating the effectiveness of test suites designed to analyze the primary model. We also propose a tree-based model to predict the error probability of the primary model for each input based on the extracted SM.
We evaluated our proposed online error prediction approach using the MNIST dataset and Mini Speech Commands dataset, achieving an area under the curve (AUC) exceeding 80\% for the receiver operating characteristic (ROC) chart.
\end{abstract}


% \begin{highlights}
% \item 
% \item 
% \item 
% \end{highlights}

% Keywords

\maketitle

% Main text
\section{Introduction}
\label{sec:introduction}

Recurrent neural network (RNN) models have made significant contributions to fields such as natural language processing and speech recognition by providing a means of processing sequential data~\cite{graves2013speech, bahdanau2014neural, lipton2015critical}. However, the lack of explainability of RNN models makes it challenging to understand their internal workings. 

Explainability, in the context of neural networks, refers to the ability to understand and interpret the decision-making process of a model, providing insights into its complex inner workings and allowing users to trust and validate the model's predictions~\cite{huang2020survey}. Explainability in RNN models is particularly challenging due to the so-called "black-box" nature of these models. The internal complexity, the non-linear transformations, and the recurrent architecture of RNNs that allow for handling of sequential data, although powerful, make it difficult to intuitively understand and interpret the decision-making process. For instance, it is not straightforward to determine which features in the input data are most influential in driving the model's predictions. Additionally, understanding the dependencies between the states of an RNN as it processes a sequence of inputs can be equally confusing due to the involved temporal dynamics and hidden state updates. This inherent lack of transparency and complexity contributes to the issue of RNNs lacking explainability. Several methods have been developed to enhance neural network explainability. Deepstellar~\cite{du2019deepstellar} automates the extraction and analysis of RNN models' internal states. Chefer et al.~\cite{chefer2021transformer} compute relevancy scores in transformer networks to clarify their decision-making. Barbiero et al.~\cite{barbiero2022entropy} use First-Order Logic to extract logic explanations from neural networks. Ayache et al.~\cite{ayache2019explaining} extract weighted automata from black box models, while the SR-RNNs approach~\cite{wang2022state} uses a "state-regularization" mechanism to improve standard RNNs. Overall, these methods allow users to gain insights into the internal mechanisms of neural network models, which in turn improves the ability to detect the potential errors in the primary neural network.

In recent literature, a significant amount of attention has been devoted to effectively testing neural networks. One of the main challenges in this area is evaluating the effectiveness of a test suite. In other words, one needs a way to quantitatively evaluate the quality of a test suite for testing a neural network. There are two main approaches for evaluating test suites in the literature; mutation and coverage criteria~\cite{huang2020survey}. Mutation testing involves making small modifications to the model, training data, or source code to reveal potential vulnerabilities and mistakes that may not be identified by conventional methods. Various approaches have been proposed for evaluating test suites of neural networks using novel mutation operators~\cite{shen2018munn, ma2018deepmutation, tambon2023probabilistic}. Coverage criteria are used to define important areas for test suite evaluation. Different coverage criteria for neural networks are introduced, including neuron coverage, condition/decision coverage, and neuron boundary coverage~\cite{pei2017deepxplore, tian2018deeptest, sun2018concolic, wicker2018feature}. These criteria have been used in various approaches for generating test cases, such as the joint optimization problem for increasing neuron coverage~\cite{pei2017deepxplore}, the greedy algorithm for testing autonomous vehicles~\cite{tian2018deeptest}, and the concolic testing approach for deep neural networks (DNNs)~\cite{sun2018concolic}. Overall, these techniques and criteria are useful for evaluating and improving the effectiveness of neural network testing.
 
However, the coverage criteria introduced in prior works, do not provide transparency into the model's decision making process, since they operate directly on the neural network \cite{huang2020survey}. Harel et al.'s research paper \cite{harel2020neuron} states that increasing neuron coverage is neither positively nor strongly correlated with improved defect detection, input realism or output impartiality. In our research, we have statistically proven that there is a relationship between the changing value of proposed coverage criteria and the error rate of the RNN model. Mutation testing strategies also require substantial computational resources for big neural networks and  may have practical limitations \cite{humbatova2021deepcrime}. This highlights a key limitation in current testing methods - the lack of coverage criteria that are both practical and provide explainability into the internal logic being tested.

In this research paper, we focus on state machine (SM) extraction from RNN-based models, which is a foundational method for modeling software systems and has significant implications for software engineering processes~\cite{lee1996principles}.
SMs provide interpretability into sequential decision processes due to their well-defined states and transitions. Our key idea is to represent the RNN model as an SM to gain insights into its inner workings. The extracted SM distills the complex RNN into a simplified representation that captures its core functionality in terms of states and transitions. Each state of the SM encompasses a cluster of similar hidden states of the RNN. The transitions between states in the SM reflect the RNN's sequence of internal state changes in response to input data. This SM representation transforms the vague RNN into an interpretable model by exposing its internal state dynamics. The conversion of the RNN into a state transition model improves model transparency by clarifying its decision-making process.
RNNs inherently possess states and state transitions, making them compatible with SM representations. By extracting an SM from an RNN-based model, we can gain insights into the model's internal workings and decision-making process. Our proposed SM extraction method differs from other existing methods \cite{du2019deepstellar,weiss2022extracting,okudono2020weighted,weiss2018practical,ayache2019explaining,weiss2019learning,wei2022extracting} in the way the states of the SM are defined. Specifically, we consider the patterns of the data and define states based on the distribution of the PM states in the state space. We also propose methods for evaluating the quality of the extracted SMs. Here, quality refers to the minimal difference in functionality between the extracted SM and the primary model (PM), which is the original RNN-based model. We compare our approach with the DeepStellar method \cite{du2019deepstellar} for extracting SM from RNN-based models, using our proposed evaluation metrics. Also, the optimal number of states for the SM extraction algorithm is determined by utilizing the proposed SM evaluation metrics.

\begin{figure}[ht]
	\centerline{\includegraphics[width=1.0\textwidth]{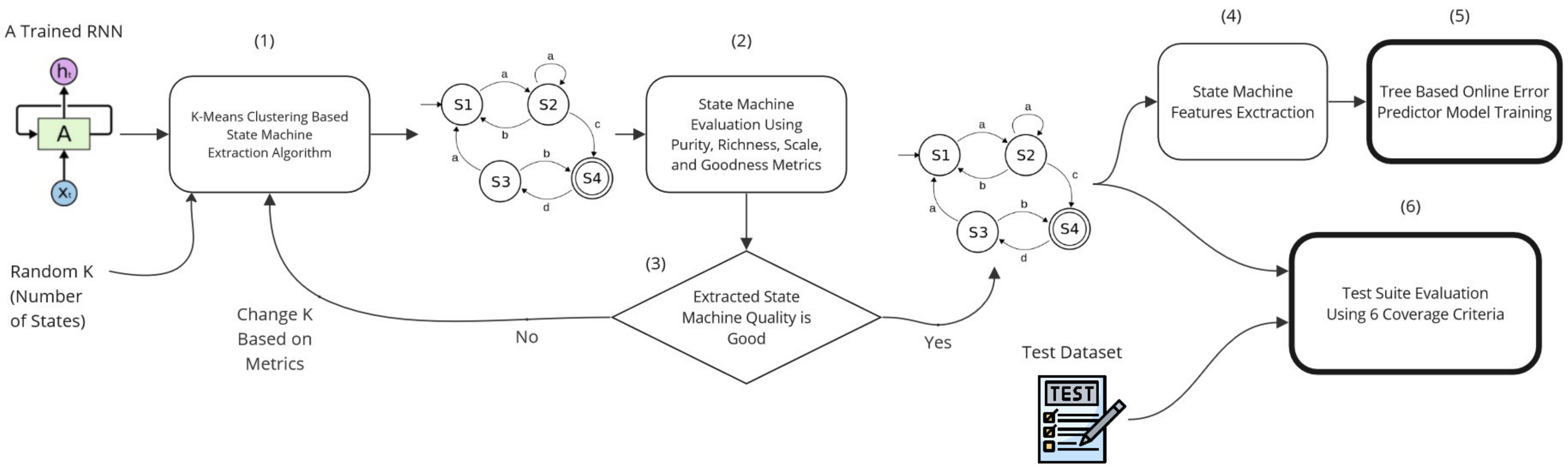}}
	\caption{DeepCover Framework}
	\label{fig:deepcover_framework}
\end{figure}

We establish coverage criteria from the extracted SM to evaluate the test suites, ensuring thorough testing of the PM's performance. By performance, we refer to the PM's accuracy, which signifies the percentage of correct predictions made by the model out of the total predictions. We also employ the extracted SM to accurately predict potential errors in the PM during the service time. The idea is to train a tree-based model based on the features derived from the state space of the extracted SM. The tree-based model can then be used to predict the error probability of the PM during the service time for each input. The step-by-step explanation of the DeepCover framework is explained in Fig. \ref{fig:deepcover_framework}. (1) The trained RNN model, designed for a critical task, is input into the framework.We employ the proposed K-Means clustering-based SM extraction algorithm to derive state machines from the RNN model. The number of states in the SM extraction algorithm is initialized randomly. (2,3) However, it is readjusted according to the proposed SM quality evaluation metrics: purity, richness, scale, and goodness. (3,4) The extracted SM, which demonstrates acceptable quality as determined by the proposed metrics, is chosen for feature extraction. This procedure is predicated on the SM states and the transitions between them. (5) Based on the extracted features and the faulty behavior of the PM, a tree-based model is trained to predict PM errors during service time. (6) Additionally, using the extracted SM and proposed coverage criteria defined on SM, any generated test suite can be evaluated in terms of considering different aspects of PM decision-making process.

To evaluate the effectiveness of our proposed method, we conducted experiments using the MNIST dataset and Mini Speech Commands dataset. Using these datasets, we trained RNN-based models that utilize GRU, LSTM, and S-RNN recurrent modules. The Mini Speech Commands dataset consists of audio recordings of simple voice commands: "yes", "no", "up", "down", "right", "left", "stop", "go". These types of commands have important applications such as controlling electronic wheelchairs or home automation systems via voice interfaces. However, misclassification of such critical voice commands can have serious consequences for users. Therefore, it is crucial that we thoroughly analyze and test the logic of speech recognition AI models like the ones trained on this dataset and uncover potential errors during the service time.
The evaluation metrics we propose demonstrate that our algorithm extract SM models of higher quality compared to those generated by previously proposed algorithms. We conduct a statistical test to assess our proposed coverage criteria, along with five coverage criteria introduced in previous studies. The statistical test results reveal a significant difference in the accuracy of the covered areas versus the overall accuracy. Furthermore, the area under the curve (AUC) of the receiver operating characteristic (ROC) chart for our implemented tree-based model surpasses 80\%, demonstrating its robust predictive performance in estimating the PM's potential errors during service time using experimental data sets.

Shortly speaking, In this research paper, we aim to answer the following research questions:
\begin{enumerate}
    \item How can we effectively extract SMs from RNN-based models, while maintaining minimal difference in functionality compared to the PM?
    \item How can we evaluate the quality of the extracted SMs?
    \item How can we use the explainability achieved through the SM extraction to improve the quality of the PM, both in testing and run-time?
\end{enumerate}
By answering these research questions, our methodology can help to improve the  explainability and dependability of RNN-based models.

Our main contributions in answering the above questions are the following:
% \begin{itemize}
% 	\item We propose an algorithm to extract an SM model from an RNN-based model. The extracted SM model serves to advance the explainability of the RNN-based model.

% 	\item Four metrics are introduced for evaluating and comparing the quality of the extracted SM models using different algorithms or configurations. 
	
% 	\item Six novel coverage criteria, based on the extracted SM, are proposed for evaluating a test suite for testing the PM. Each coverage criterion is focused on a specific aspect of the SM and its state space.
	
% 	\item We further propose a tree-based model that leverages coverage features derived from the extracted SM, aiming to predict potential errors in the primary model during its service time. 
% \end{itemize}

\begin{itemize}

\item We propose a K-Means clustering based algorithm to extract an SM model from an RNN-based model. The algorithm considers the distribution of PM states and defines SM states accordingly, helping to advance the explainability of the PM.

\item Four metrics are introduced to evaluate the quality of extracted SM models:
\begin{itemize}
\item Purity: Evaluates SM quality by comparing PM final state labels
\item Richness: Assesses SM quality based on PM final states per SM state
\item Goodness: Combines Purity and Richness into a single metric
\item Scale: Measures SM discrimination ability relative to labels
\end{itemize}

\item Six coverage criteria focused on specific aspects of the SM and its state space are proposed:
\begin{itemize}
\item New Final State Coverage
\item Out of Boundary Final State Coverage
\item Basic Final State Coverage
\item Basic Label and Final State Coverage
\item Weighted Basic Label and Final State Coverage
\item Weighted Label and Final State Coverage
\end{itemize}
The efficacy of criteria is validated using statistical Kolmogorov-Smirnov testing \cite{kolmogorov1933sulla}, comparing covered areas accuracy vs overall accuracy.

\item We propose a tree-based model leveraging explainable features from the SM state space to predict potential errors in the PM during its service time. Our experiments demonstrate over 80\% AUC-ROC in error prediction.

\end{itemize}

The paper is structured as follows. In  Section~\ref{sec:related}, we review the existing literature on explainability of RNNs and test suite evaluation using coverage criteria and mutation. Section~\ref{sec:methodology} presents the proposed methods for extracting an SM, evaluating metrics, and coverage criteria, as well as online error prediction method. In Section~\ref{sec:experiments}, we present how the proposed methodologies are evaluated using datasets and RNN-based models, including GRU, LSTM, and S-RNN. Finally, Section~\ref{sec:conclusion} summarizes the paper's contributions and findings, and provides suggestions for future research directions. 

\section{Related Work} 
\label{sec:related}

In this section, we briefly present the literature on RNN explainability, neural network test suite evaluation, and online error prediction. 

\subsection{Explainability}
The field of neural networks greatly values the concept of explainability, as it provides a means to enhance comprehension and gain deeper insights into the inner workings of these complex models \cite{huang2020survey}. It can also be used to improve the accuracy of neural networks.

A common category of research in this field focuses on the extraction of automaton-like structures from neural networks. Weiss et al. \cite{weiss2022extracting} introduced a method to extract a deterministic finite automaton (DFA) from a trained S-RNN using Angluin's L* algorithm~\cite{Angluin1987}. The DFA represents the state dynamics of the RNN, using the trained RNN as an oracle. Okudono et al.~\cite{okudono2020weighted}, Ayache et al. \cite{ayache2019explaining}, and Wei et al. \cite{wei2022extracting} proposed similar approaches for extracting weighted finite automaton (WFA) from RNNs, which help in improving interpretability and reducing inference cost. Weiss et al. \cite{weiss2019learning} further extended the automata extraction concept by proposing a method to extract a Probabilistic Deterministic Finite Automaton (PDFA) from black-box RNN language models.
According to Ayache et al.~\cite{ayache2019explaining}, DFAs, WFAs, and PDFAs may not accurately represent or mimic the behavior of LSTMs due to disparities in their language coverage capabilities. In a similar vein, William Merrill \cite{merrill2019sequential} suggested that RNNs, including LSTMs, function like counter machines, thus making them more powerful than regular languages. He enhanced neural network interpretability by connecting them to automata and formal language theory, and introduced asymptotic acceptance to characterize various recurrent neural networks.

Another set of studies focus on deep understanding and decomposition of neural networks. Dosovitskiy et al.~\cite{dosovitskiy2016inverting} proposed a technique using an up-convolutional neural network to analyze feature representations from visual data. Chefer et al.~\cite{chefer2021transformer} proposed a method for calculating relevancy scores in transformer networks, enabling better understanding of their decision-making process. They introduce Deep Taylor Decomposition to maintain total relevance across transformer network layers, including attention and skip connections. This approach targets transformer networks by identifying influential feature vectors without explaining inner mechanisms

Stateful NNs are a class of neural networks that maintain an internal state, which is updated with each input processed, allowing them to model temporal dependencies and handle sequential data. RNNs are a typical example of stateful NNs. In terms of offering explainability through these stateful NNs, Du et al.~\cite{du2019deepstellar} introduced Deepstellar, a framework that provides an automated approach to extract and analyze the internal states of deep learning models. It converts the state space of RNN-based models into a Discrete-Time Markov Chain (DTMC) to explain the model's inner function. Discretizing state space with static gridding and reducing dimensions using PCA may not optimally explain RNN models' functionality due to pattern loss. The State-Regularized Recurrent Neural Networks (SR-RNNs) approach by Wang et al.~\cite{wang2022state} also enhances the interpretability and explainability of RNNs by using a stochastic state transition mechanism called "state-regularization". 

In the context of logic-based explainability, Barbiero et al. \cite{barbiero2022entropy} presented an end-to-end differentiable method to extract logic explanations from neural networks using First-Order Logic. The entropy-based criterion identifies relevant concepts, enabling succinct explanations in safety-critical domains like clinical data and computer vision.

\subsection{Test Suite Evaluation}
This subsection presents various approaches for evaluating test suites for neural networks in two cateogries of mutation and coverage criteria.

\subsubsection{Mutation}
Mutation testing stands as one of the most robust methods for test design. The underlying principle of mutation testing involves modifying the source code of a program by making small syntactic alterations, referred to as mutation operators. These language-specific mutation operators are carefully crafted, taking into account the characteristics of the language or the common mistakes made by software developers~\cite{huang2020survey}. 

Numerous research studies have tried to implement the mutation testing approach within the realm of neural networks by introduction new mutation operators for this domain. Shen et al.~\cite{shen2018munn} proposed five mutation operators targeting input neurons, hidden layer neurons, activation functions, biases, and weights. 
In~\cite{ma2018deepmutation}, different mutation operators targeting neural network source code, training data, and architecture are introduced. 

Tambon et al.~\cite{tambon2023probabilistic} introduced Probabilistic Mutation Testing (PMT) for DNNs, considering the stochasticity during training. PMT provides consistent decisions on mutant killing and demonstrated effectiveness using three models and eight mutation operators.

\subsubsection{Coverage Criteria}
Other research in the field of test suite evaluation and generation has focused on defining coverage criteria. In the realm of software testing, coverage criteria serve as metrics that measure the extent of testing carried out by a specific set of tests.  These criteria can identify areas that need additional testing~\cite{huang2020survey}. They can also be used as a guideline for generation of effective test cases.

Numerous efforts have been made to tackle the definition of coverage criteria within the context of testing neural networks. Neuron Coverage (NC) quantifies the percentage of activated neurons in a neural network, with the implicit assumption that enhancing NC leads to an enhancement in the quality of a test suite~\cite{pei2017deepxplore}. Within a similar context, DeepTest systematically examines various segments of the deep neural network logic by generating test inputs that maximize the count of activated neurons~\cite{tian2018deeptest}.
 In~\cite{sun2018concolic}, neuron coverage, condition/decision coverage, and neuron boundary coverage criteria are used as test requirements to generate test cases and add them to the test suite. Similarly, Ma et al.~\cite{ma2018deepgauge} proposed a neural network testing approach using two coverage criteria of neuron-level and layer-level. 

Inspired by input space partitioning approach in software testing, Wicker et al.~\cite{wicker2018feature} proposed to discretize the input space into hyper rectangles with the goal of generating test cases that cover the entire input space.
Kim et al.~\cite{kim2019guiding} proposed a surprise adequacy-based approach for testing NNs, introducing unexpected inputs and measuring the difference between expected and actual outputs. They defined Surprise Coverage to guide testing and proposed two methods for measuring surprise adequacy.

Deepstellar~\cite{du2019deepstellar} extracts a DTMC model from RNN-based models for explainability and defines coverage criteria for test suite evaluation. The criteria include: 1) Basic State Coverage, covering visited DTMC states; 2) Weighted State Coverage, covering rarely met states; 3) n-Step State Boundary Coverage, covering states on the primary model's margin; 4) Basic Transition Coverage, covering visited DTMC transitions; and 5) Weighted Transition Coverage, covering infrequent transitions. These criteria serve as an objective function for generating new test cases.

\subsection{Online Error Prediction}
 % Online error prediction in neural networks is crucial in safety-critical applications like autonomous vehicles, where early error detection can prevent accidents and enhance system reliability~\cite{huang2020survey}. 

By online error prediction, we refer to predicting the error probability of the primary model for each input in real-time, as new input data is processed.

Ross et al.~\cite{ross2018improving} present a defense method using input gradient regularization, enhancing neural network robustness against adversarial examples. Rossolini et al.~\cite{rossolini2022increasing} proposed an approach with similar goal using coverage analysis. By computing a confidence value based on the activation state of internal neurons and comparing it to a trusted dataset, the method effectively identifies adversarial examples  with minimal impact on execution time and memory requirements.

Another approach for detecting adversarial examples is presented in~\cite{hou2022similarity}, where the authors proposed a similarity-based integrity protection method (IPDLS) for deep learning systems. By measuring similarity between suspicious samples and a preset verification set, IPDLS performs anomaly detection.
Raghunathan et al.~\cite{raghunathan2018certified} proposed a certified defense against adversarial examples for single-hidden-layer neural networks. They developed a semidefinite relaxation-based method that outputs a differentiable certificate, ensuring no attack can force the error beyond a threshold. This enables joint optimization with network parameters and robustness against all attacks. 

Antoran et al.~\cite{antoran2020depth} present a method that estimates model uncertainty in neural networks by performing probabilistic reasoning over different depths, corresponding to subnetworks. Exploiting the sequential structure of feed-forward networks allows for single-pass evaluation and prediction, offering uncertainty calibration, robustness to dataset shifts, and competitive accuracies with less computational cost.

In summary, many works focus on CNNs instead of RNNs, and few coverage criteria use explainability or provide a clear representation of the primary model's decision-making process. Most works evaluate test suites without extracting interpretable models from RNNs. Language coverage limitations of extracted explainable models, such as DFA, PDFA, and WFA, are also important, as they struggle to mimic RNNs' language coverage, especially LSTMs \cite{merrill2019sequential, ayache2019explaining}.

Current research on neural network explainability and test suite evaluation has advanced our understanding of their performance. However, a comprehensive method targeting RNNs, extracting interpretable models like state machines, and defining coverage criteria based on these models is still needed. Our methodology centers on extracting a state machine from RNN-based models to enhance explainability, enabling coverage criteria definition for test suite evaluation and online error prediction. This approach improves understanding of RNNs' internal decision-making and assists in identifying potential errors in the primary model at the design and service time.

\section{Methodology}
\label{sec:methodology}
In this section, we offer a thorough technical description of the proposed methods. First, we provide an overview of RNNs and their various types, which form the foundation for the models employed in this research. Following that, we detail our approach for extracting a state machine from RNN models. We also propose four metrics to assess the quality of the extracted SMs. Furthermore, we introduce six coverage criteria based on the state space of SMs to evaluate the effectiveness of test suites. Finally, we present a method for online error prediction of the primary model by training a tree-based model using the state space of the derived SM. 

\subsection{Recurrent Neural Networks}
\label{recurrent_neural_networks_subsection}
RNNs are a class of neural networks designed for handling sequential inputs and outputs. They are widely used in various problem domains and come in different types, including Simple Recurrent Neural Networks (S-RNNs)~\cite{elman1990finding}, Long Short Term Memory (LSTM) networks~\cite{hochreiter1997long}, and Gated Recurrent Unit (GRU) networks~\cite{bahdanau2014neural}.

% All RNNs process sequential inputs and generate sequential or single-vector outputs. At each time step, an input vector is provided to the RNN module, which generates a state vector based on its internal function. The output vector generated at each time step may be identical to the state vector, as is the case for S-RNNs and GRUs, or it may be the result of a different function, as is the case for LSTMs.

\begin{figure}[ht]
	\centerline{\includegraphics[width=0.5\textwidth]{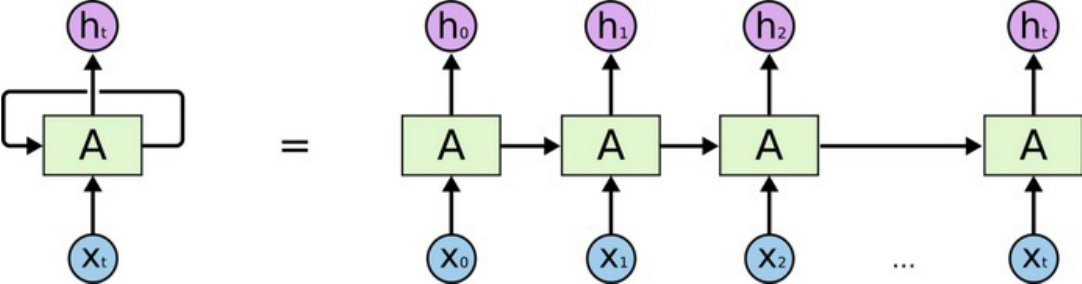}}
	\caption{General Function Schema of an RNN Module}
	\label{fig:rnn_fig}
\end{figure}

As illustrated in Fig. \ref{fig:rnn_fig}, all RNNs share certain fundamental characteristics. They process sequential inputs, and their outputs can also be sequential or a single vector representing the final output. At each time step, an input vector $x_{t}$ is provided to the RNN module, which generates a state vector $h_{t}$ based on its internal function. The output vector generated at each time step may be identical to the state vector, as is the case for S-RNNs and GRUs, or it may be the result of a different function, as is the case for LSTMs. This output vector is then used as an input for the next time step, in addition to the original input vector $x_{t}$.

\subsubsection{Simple Recurrent Neural Network}
The S-RNN is a type of RNN  \cite{elman1990finding} that is considered to be the oldest and simplest form of RNNs. According to eq. \ref{SRNN_introduction_eq}, at each time step, the input vector $x_t$ is combined with the previous time step's state vector. However, this simple function is not able to effectively extract features from long input sequences or maintain a memory of the earliest extracted features.

\begin{figure}[ht]
	\centerline{\includegraphics[width=0.3\textwidth]{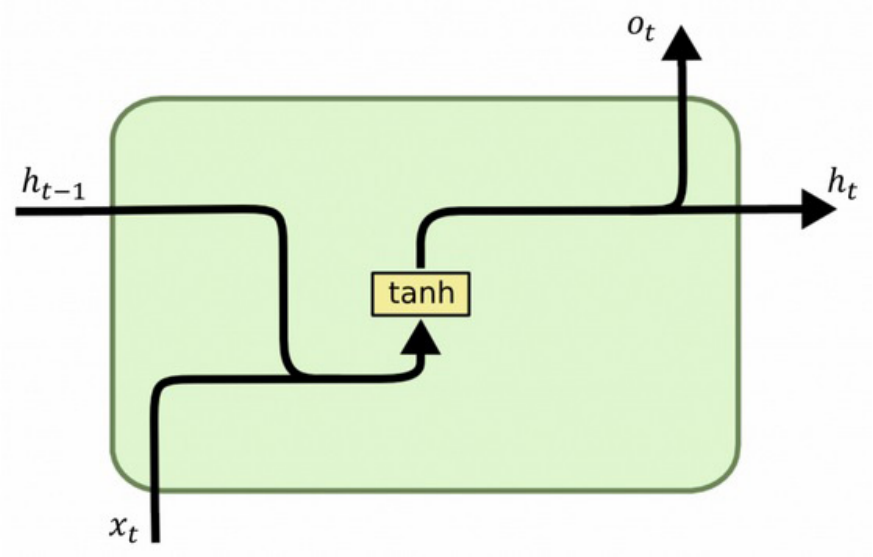}}
	\caption{Internal Architecture of an S-RNN}
	\label{fig:recurrent_neural_networks}
\end{figure}

\begin{equation}\label{SRNN_introduction_eq}
    h_{t} = tanh(Wx_{t} + Uh_{t-1} + b)
\end{equation}

\subsubsection{Long Short Term Memory}
\begin{figure}[ht]
	\centerline{\includegraphics[width=0.3\textwidth]{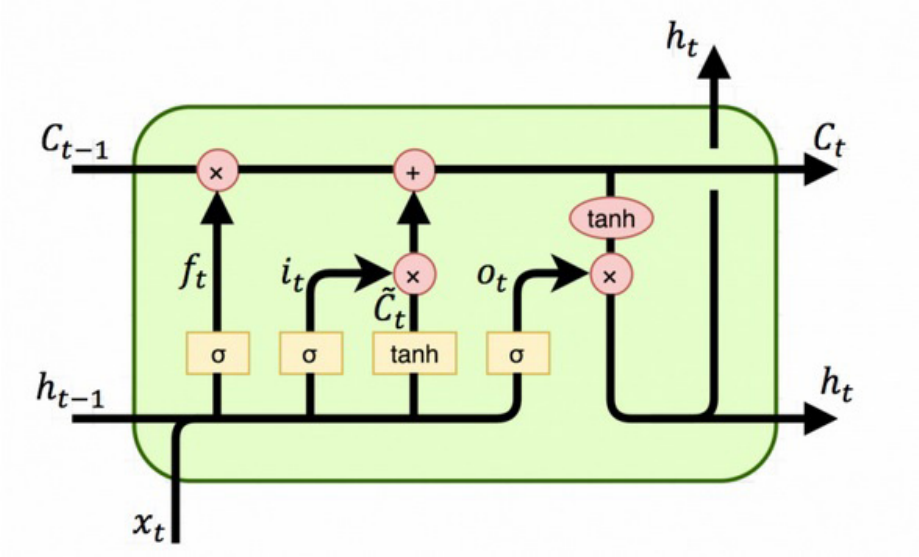}}
	\caption{Internal Architecture of an LSTM Module}
	\label{fig:long_short_term_memory}
\end{figure}

In contrast, the LSTM module, introduced in 1997, was specifically designed to address the S-RNN's issue of "Gradient Vanishing" on long input sequences. The internal function of the LSTM (Fig. \ref{fig:long_short_term_memory}) is defined by eq. \ref{lstm_intro_1_eq} to \ref{lstm_intro_6_eq}, where the operator $\odot$ is the Hadamard Product, which results in the element-wise product of two matrices. The LSTM module contains an information band ($c_t$) which, at each time step, is used to write new information and manipulate older information. This information band enables the LSTM to effectively extract features from long input sequences while maintaining a memory of earlier inputs. The output at each time step is the information band vector ($c_t$) in addition to the state vector ($h_t$).

Based on eq. \ref{lstm_intro_1_eq}, \ref{lstm_intro_2_eq} and \ref{lstm_intro_3_eq}, at each time step the input vector ($x_t$) is combined with the previous time step's state vector ($h_t$) using three different weights. New information for the information band is generated as defined in eq. \ref{lstm_intro_4_eq}, and then combined with old information on the information band based on the combined weight (eq. \ref{lstm_intro_5_eq}). The current time step's state vector ($h_t$) is generated based on the current input vector ($x_t$), previous time step's state vector ($h_{t-1}$) and new information vector ($c_t$) for the information band, as defined in eq. \ref{lstm_intro_6_eq}. All of these weights are optimized during the training phase of the model.
 
\begin{equation}\label{lstm_intro_1_eq}
    f_{t} = \sigma(W^{f}x_{t} + U^{f}h_{t-1} + b^{f})
\end{equation}
\begin{equation}\label{lstm_intro_2_eq}
    i_{t} = \sigma(W^{i}x_{t} + U^{i}h_{t-1} + b^{i})
\end{equation}
\begin{equation}\label{lstm_intro_3_eq}
    o_{t} = \sigma(W^{o}x_{t} + U^{o}h_{t-1} + b^{o})
\end{equation}
\begin{equation}\label{lstm_intro_4_eq}
    \widetilde{c}_{t} = tanh(W^{c}x_{t} + U^{c}h_{t-1} + b^{c})
\end{equation}
\begin{equation}\label{lstm_intro_5_eq}
    c_{t} = f_{t} \odot c_{t-1} + i_{t} \odot \widetilde{c}_{t}
\end{equation}
\begin{equation}\label{lstm_intro_6_eq}
    h_{t} = o_{t} \odot tanh(c_{t})
\end{equation}

\subsubsection{Gated Recurrent Unit}
Despite the success of the LSTM model, its training time is often too slow and its resource consumption is high. In order to address this issue, the GRU \cite{bahdanau2014neural} was introduced in 2014. The key difference between the LSTM and the GRU is that while the LSTM can write and delete information on its information band at each time step, the GRU only performs one of these actions. This limitation results in lower resource consumption during the training phase of GRU modules.

\begin{figure}[ht]
	\centerline{\includegraphics[width=0.3\textwidth]{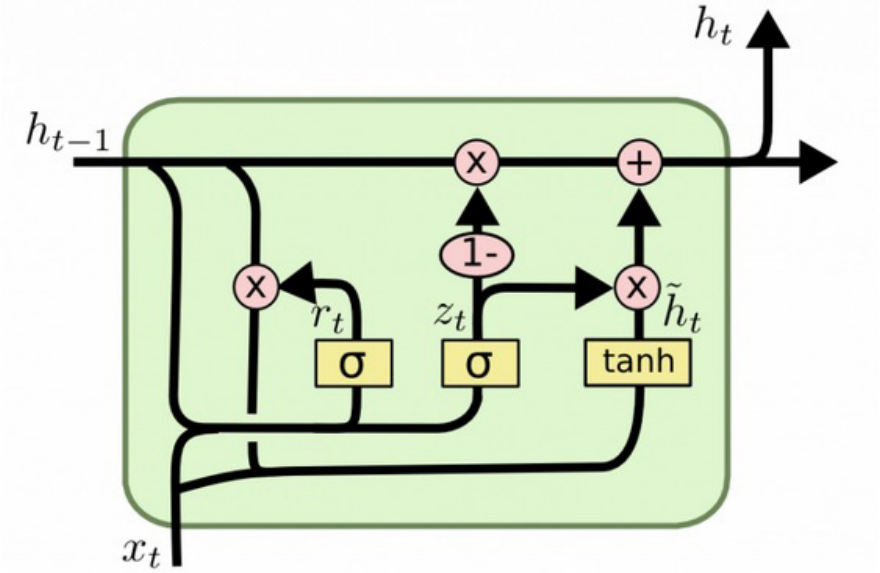}}
	\caption{Internal Architecture of a GRU Module}
	\label{fig:gated_recurrent_unit}
\end{figure}

The internal structure of the GRU is illustrated in Fig. \ref{fig:gated_recurrent_unit} and is defined by eq. \ref{gru_intro_1_eq} to \ref{gru_intro_4_eq}. As outlined in eq. \ref{gru_intro_1_eq} and eq. \ref{gru_intro_2_eq}, the input vector of the current time step is combined with the state vector of the previous time step using two different weights. Subsequently, as defined in eq. \ref{gru_intro_3_eq}, new information is generated to be appended to the information band. Finally, as outlined in eq. \ref{gru_intro_4_eq}, the GRU writes the latest information on the information band or preserves the older information.

\begin{equation}\label{gru_intro_1_eq}
    z_{t} = \sigma(W^{z}x_{t} + U^{z}h_{t-1} + b^{z})
\end{equation}
\begin{equation}\label{gru_intro_2_eq}
    r_{t} = \sigma(W^{r}x_{t} + U^{r}h_{t-1} + b^{r})
\end{equation}
\begin{equation}\label{gru_intro_3_eq}
    u_{t} = tanh(W^{u}x_{t} + U^{u}(r_{t} \odot h_{t-1}) + b^{u})
\end{equation}
\begin{equation}\label{gru_intro_4_eq}
    h_{t} = z_{t} \odot h_{t-1} + (1-z_{t}) \odot u_{t}
\end{equation}

% S-RNNs are the oldest and simplest form of RNNs, but they struggle to extract features from long input sequences or maintain a memory of the earliest extracted features. LSTMs were introduced to address this issue by incorporating an information band that allows the network to effectively extract features from long input sequences while maintaining a memory of earlier inputs. However, LSTMs can be slow to train and resource-intensive.

% GRUs were introduced to address the training time and resource consumption issues of LSTMs. The main difference between LSTMs and GRUs is that while LSTMs can write and delete information on their information band at each time step, GRUs only perform one of these actions, resulting in lower resource consumption during training.

\subsection{State Machine Extraction}
\label{sec:extraction}
To test an RNN model, we propose an algorithm to extract an SM as an explainable model from the primary model. Our methodology for SM extraction is inspired by the Deepstellar approach \cite{du2019deepstellar}. However, the Deepstellar method discretizes the state's space using static gridding and reduces dimensions through PCA. We argue that such an approach may not optimally capture the functional behavior of RNN models due to potential loss of intricate patterns. Consequently, we propose an enhanced technique for extracting an SM from RNN-based models that provides a more representative view of the internal decision-making process in these models. The following sections elaborate on the detailed implementation of our approach. We can define the coverage criteria for test suite evaluation and implement an error prediction model with the help of transparency  provided by the derived SM.

We have chosen to use state machines, since first, state machines provide a clear and structured representation of the internal decision-making processes in RNN models, making them more explainable and interpretable. Secondly, state machines are well-suited for modeling stateful NNs, such as RNNs, due to their inherent ability to handle sequential data and temporal dependencies. This makes state machines a natural choice for representing stateful NNs and analyzing their inner workings.

An SM is composed of a set of states and transitions between those states. At each time step, the SM generates output upon transitioning to a new state after receiving input. SMs have been extensively utilized to model systems in a variety of fields, including sequential circuits, programs, communication protocols, and machine learning models, as demonstrated in the literature~\cite{friedman1971fault,kohavi2009switching,weiss2022extracting}.
 More formally, a simple SM can be defined as follows:
\begin{equation}\label{fsm_eq}
M = (I, O, S, \delta, \lambda, F)
\end{equation}
, where I, O, S, and F represent non-empty sets of input symbols, output symbols, states, and final states respectively. Final states are a subset of states that determine the acceptance of an input string. The transition function $(\delta)$ determines the next state based on the current state and current time step input as defined in Equation~\ref{state_transition_eq}. The output function $(\lambda)$ generates the SM output based on the current state and time step input as shown in Equation~\ref{output_function_eq}.

\begin{equation}\label{state_transition_eq}
\delta: S\times I \rightarrow S
\end{equation}
\begin{equation}\label{output_function_eq}
\lambda: S\times I \rightarrow O
\end{equation}

\begin{figure}[ht]
	\centerline{\includegraphics[width=0.7\textwidth]{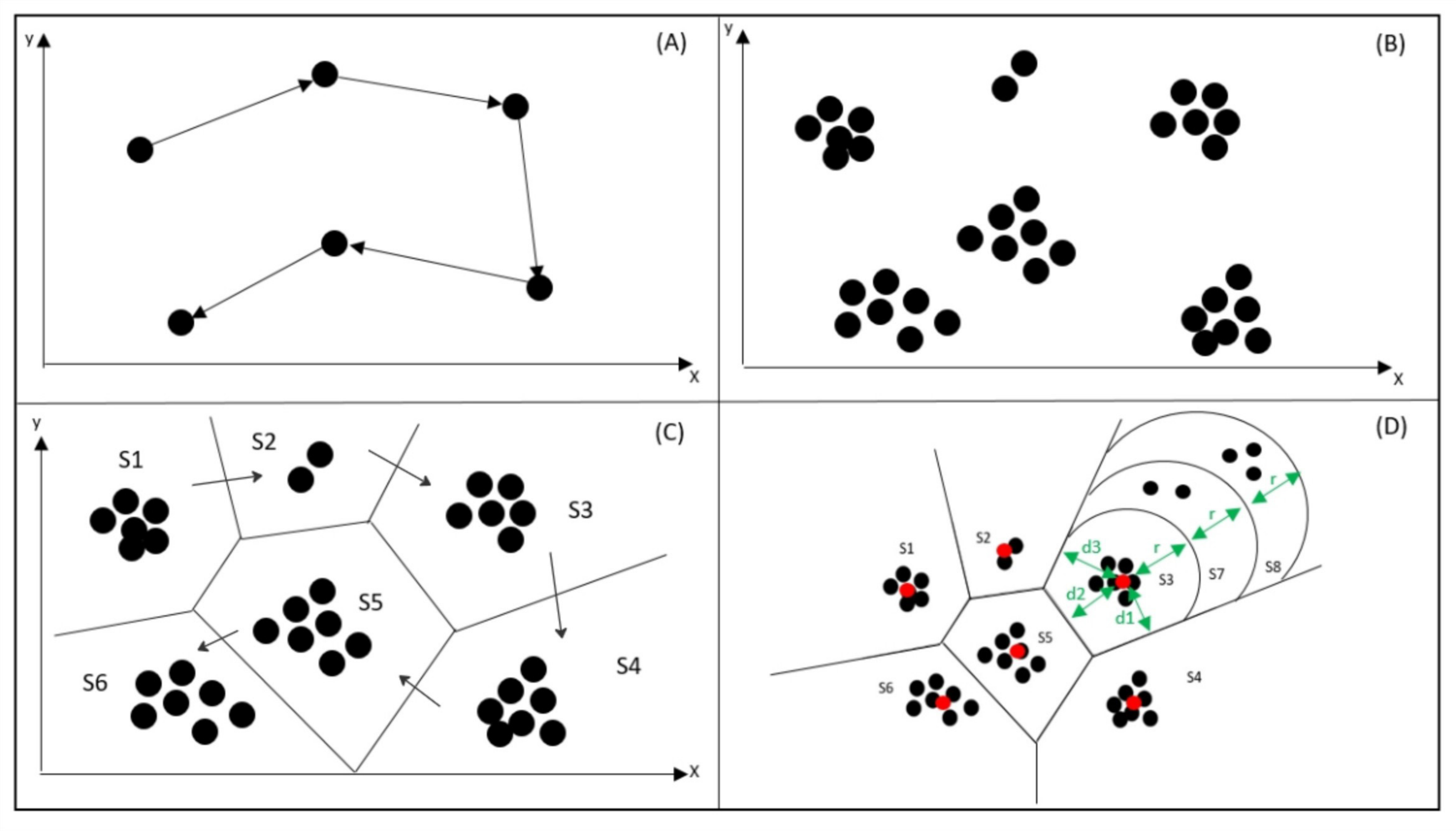}}
	\caption{
		(A) A representation of the trace of states in a two-dimensional state space, as a result of a single input sequence of length six. 
		(B) A visualization of multiple traces of states in a two-dimensional state space, resulting from multiple input sequences. 
		(C) The clustering of training data state vectors using the K-Means clustering algorithm, with each cluster being considered as a state in the extracted SM. 
		(D) The identification of new states for the PM, for instances where the PM states are farther from their respective centroids than what is deemed acceptable.
	}
	\label{fig:sm_extraction_steps}
\end{figure}

As discussed in Section~\ref{recurrent_neural_networks_subsection} on RNNs, the state vector at each time step is returned by RNN modules for a given input sequence. As illustrated in Fig.~\ref{fig:sm_extraction_steps}.A, each state vector can be considered as a point in the state space, and the input sequence results in a trace of states, depicting the movement in the state space. In order to define the state space of the PM, all the Training Dataset (TD) is inputted to the model after the training phase. The state vectors are extracted for all inputs, as depicted in Fig.~\ref{fig:sm_extraction_steps}.B. For the purpose of this study, we denote the set of PM state vectors as PMS, and the maximum time step as MTS. Then the PMS can be defined as follows:

\begin{equation}\label{primary_model_states_set_eq}
	PMS = \{ PM(td_{i}).states[j]  \: | \: 1\leq i \leq \|TD\| , 1 \leq j \leq MTS , td_i \in TD \}
\end{equation} 

In order to extract an SM from an RNN-based model, it is necessary to discretize the state space to define the SM states. This process is depicted in Fig.~\ref{fig:sm_extraction_steps}.C. We apply the \textit{K-Means clustering algorithm} to the set of PMS to extract the SM states. K-Means is a  clustering algorithm, where data points are separated into pre-specified number of groups by minimizing a criterion known as Inertia~\cite{vassilvitskii2006k}. Each cluster is represented by means of the samples in that cluster, commonly referred to as the cluster centroid. The K-Means algorithm is iterative, continuously attempting to minimize the Inertia and reposition the centroids to the optimal locations.

After the state machine's states are established during the extraction process, the state machine can be utilized for coverage criteria and real-time error prediction with test data. It is crucial to note that the test data might enter states that are not included in any of the states of the extracted state machine. As a result, it is essential to have a method for defining new states while the state machine is being used with test data. We set a maximum distance limitation for new PM states to the centroids in marginal clusters to define new SM states. This is illustrated in Fig.~\ref{fig:sm_extraction_steps}.D, where a maximum distance limitation, denoted as $r$, is applied to the centroids for new PM states in marginal clusters. We represent the marginal cluster centroid distance to its  $i$th neighbor border as $d_i$, and the number of neighbor clusters as $NC$. The maximum distance limitation can be defined mathematically as $r=Max(\{d_i | 1 \leq i \leq NC \})$. In the case that a new PM state is located in a marginal cluster and its distance to the centroid exceeds $r$, a new state for the SM is defined, with a width equivalent to $r$.

Having the set of states and following Equation~\ref{fsm_eq}, we define the extracted SM from the PM as follows:

\begin{equation}\label{state_machine_extraction_eq}
SM = (I, O, SMS, \delta, \lambda, SMFS)
\end{equation}

\noindent , where  $I$ and $O$ represent the sets of input and output vectors, respectively. $SMS$ represents the SM states and $SMFS$ represents the final states in the state machine. The functions $\delta$ and $\lambda$ are defined based on the PM. In other words, to generate the next state vector and the next output vector at each time step, the RNN-based model should be utilized. Based on these, we can generate the next state vector and the output vector for each input by utilizing the $\delta$ and $\lambda$ corresponding to the transition function and output function of the primary RNN model. An illustration of this process is provided in Fig.~\ref{fig:sm_extraction_steps}.A. This figure demonstrates how transitions between primary model states dictate the transitions between states within the SM, as further depicted in Fig.~\ref{fig:sm_extraction_steps}.C. The output vector is constructed through the utilization of the primary model; $\lambda$ symbolizes this specific component of the primary model.

The final states are determined based on the last time step of inputs on the RNN-based model  (the state on which an input is finished). In other words, $SMFS$ consists of those states from $SMS$ where at least one final state of the primary model is present.

In terms of computational complexity, the proposed SM extraction approach involves gathering the state vectors for the entire training dataset, followed by clustering them using K-Means to define the states. Gathering the full set of state vectors takes $O(\|TD\| * MTS * RNNT)$ time where $\|TD\|$ is the size of the training set, $MTS$ is the maximum sequence length, and $RNNT$ is the time for RNN inference per input. The K-Means clustering algorithm itself has an average case complexity of $O( (\|TD\| * MTS) * \|SMS\| * I)$ where $\|TD\| * MTS$ is the number of data points, $\|SMS\|$ is the number of clusters, and $I$ is the number of iterations \cite{vassilvitskii2006k}. Since $\|SMS\|$ and $I$ are constants defined a priori, the overall time complexity is linear in terms of the total number of state vectors extracted.

\subsection{State Machine Evaluation Metrics}

The literature is missing metrics for evaluating the quality of the state machines extracted from the neural networks. The metrics can be used to determine the optimal hyperparameters of the algorithms, like the number of centroids in the K-Means clustering algorithm in our case. Also, using these metrics, different algorithms for SM extraction can be compared. In this section, we present 4 metrics for evaluating SMs extracted from RNNs.

\subsubsection{Purity}
The Purity metric is used to evaluate the quality of the SM extracted from the PM. It does this by comparing the labels of the PM's final states within each state of the SM. If there is a mismatch in the labels, it indicates that the SM extraction algorithm has failed to distinguish between different types of PM final states.

The Purity metric is calculated by examining the PM final states with different predicted labels $(pl)$ within the same SM state. The more PM final states with different predicted labels in each SM state, the lower the Purity of the extracted SM. Mathematically, the Purity metric is defined as follows:

\begin{equation}\label{purity_eq}
Purity = \frac{\sum_{i=1}^{\|SMS\|} max(\{  \| \{  s | s.pl = l , s \in SMS[i] \} \| \; | \; l \in L \})}{\sum_{i=1}^{\|SMS\|} \| \{s \; | \; s \in SMS[i] , s.pl \neq Null \} \|}
\end{equation}

\noindent, where $L$ represents the set of class labels. The Purity metric calculates the ratio of the total number of major PM final states (biggest set of final states with the same predicted label) in each state of the SM to all PM final states. The predicted label for non-final states is Null.

For example, different derived state machines are depicted in  Fig.~\ref{fig:sm_evaluation_metrics}. The black states correspond to non-final states of the primary RNN model,  while the blue and green colored states represent the final states, with each color indicating a separate predicted label. Now, for SM (A), the maximum number of states with the same label for the states are 0, 1, 1, 1, 1, 1 for the states S1-S6, since in the first SM state, there is no final state, while in the rest, the maximum number of final states with the same label is 1. Since the number of all final states  is 7, the Purity for SM (A) equals:

\begin{equation*}
\frac{0+1+1+1+1+1}{7} = \frac{5}{7}
\end{equation*}

Similarly, the Purity of SM (B) can be calculated as follows:

\begin{equation*}
    \frac{0+1+1+2+1+2}{7} = \frac{7}{7}
\end{equation*}

\noindent, and hence, the Purity of SM (B) is considered to be higher than that of SM (A). Intuitively, that is due to the fact that in SM (B), there are no PM final states with varying class labels in one SM state.

\subsubsection{Richness}

If every final state of the primary model resides in one distinct state of the extracted SM, then the Purity of the SM equals one. However, to mimic the primary RNN model effectively, it is preferable to group the final states with the same label under the same SM state. The Richness metric is defined to assess the extracted SM's quality based on this feature. It calculates the average number of PM final states in each SM state. More formally, the Richness metric is defined as follows.

\begin{equation}\label{richness_eq}
Richness = \frac{\sum_{i=1}^{\|SMS\|} \| \{s \; | \; s \in SMS[i] , s.pl \neq Null \} \|}{\sum_{i=1}^{\|SMS\|} \| \{i \; | \; \exists s \in SMS[i] : s.pl \neq Null \} \|}
\end{equation}

The number of PM final states in each SM state is in the numerator, and the number of SM states, including PM final states, is in the fraction's denominator. As visualized in Fig.~\ref{fig:sm_evaluation_metrics}, the SM (C) has more Richness than SM (B), since the PM's final states are distributed in fewer SM states.

For example, let's consider the same state machines in Fig.~\ref{fig:sm_evaluation_metrics} mentioned above. The number of final states in each state of SM (B) are 0, 1, 1, 2, 1, 2 for states S1-S6 and the number of SM states that include PM final states is 5. Thus, the Richness for SM (B) would be calculated as:

\begin{equation*}
    \frac{0+1+1+2+1+2}{5} = \frac{7}{5}
\end{equation*}

In a similar manner, we can calculate Richness for SM (C) as follows:

\begin{equation*}
    \frac{0+1+0+0+2+4}{3} = \frac{7}{3}
\end{equation*}

So, according to the Richness metric, SM (C) is richer than SM (B) because it is more effective in grouping the PM final states with the same label under the same extracted SM state. The more rich an extracted state machine is, the better it can reflect the behavior of the primary RNN model.

\subsubsection{Goodness}
The goodness metric combines the Purity and Richness metrics, making it easier to evaluate the extracted SM based on these two requirements. According to Equation~\ref{goodness_eq}, the Purity to the power of 10 helps to increase the importance of high Purity in an extracted SM, and it means that high Richness can not affect the Goodness score without the help of a good Purity score. The value of the power can be changed based on the sensitivity of the problem to the Purity metric. The Purity raised to the power of 10 means a Purity score lower than 50\% is unacceptable. This happens because when the Purity score is below 0.5, raising it to the 10th power results in a Goodness score close to zero, regardless of the Richness value.

\begin{equation}\label{goodness_eq}
Goodness = Purity^{10} * Richness
\end{equation}

\subsubsection{Scale}
The other measure to evaluate the extracted SM is the number of derived states. The number of states in the SM should be at least the number of labels in the primary model, since the labels need to be discriminated. However, if the number of states in the SM is much more than the number of class labels, this indicates unnecessary complexity in the model, which leads to less explainability of the derived SM.

\begin{figure}[ht]
	\centerline{\includegraphics[width=0.7\textwidth]{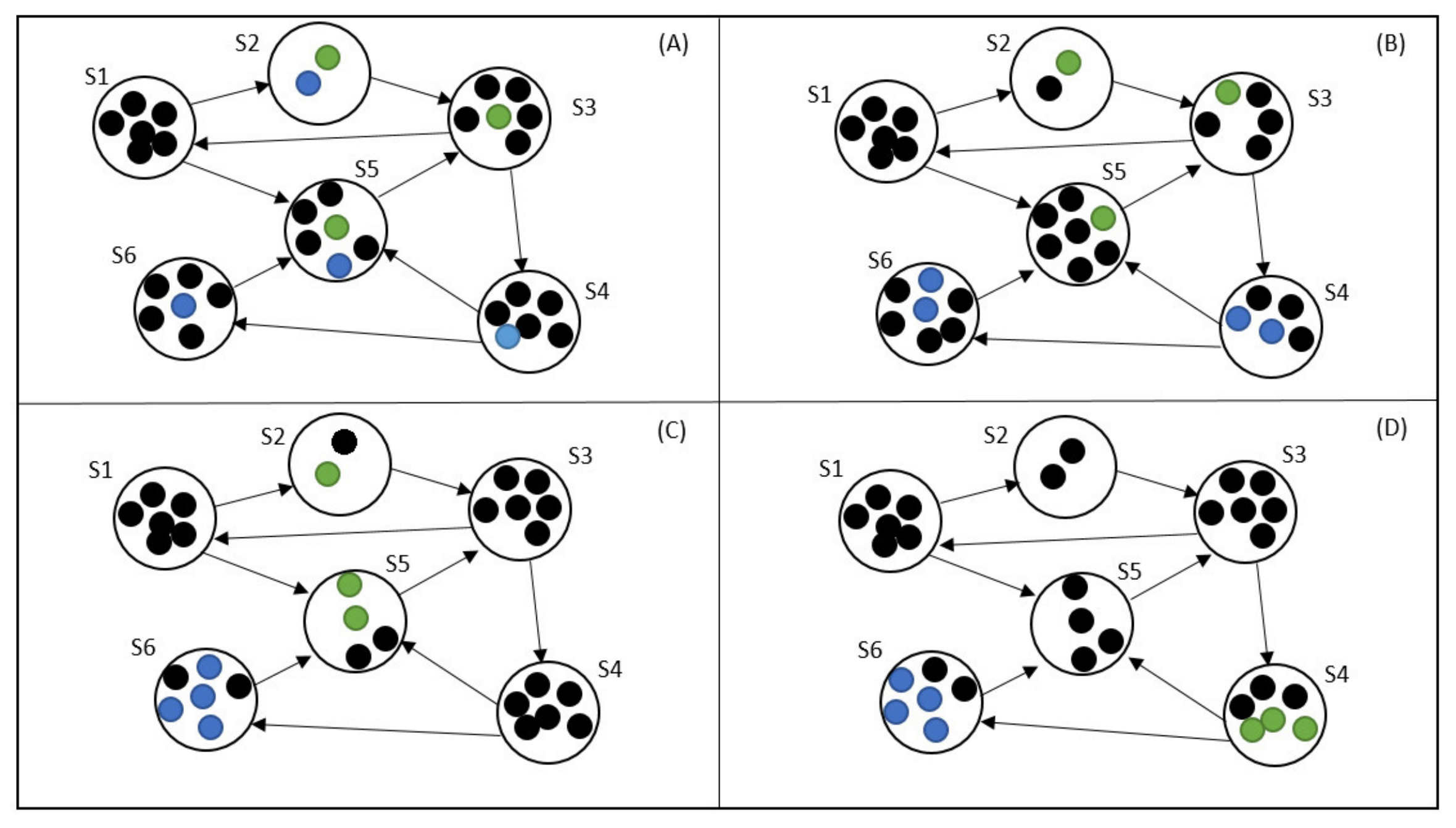}}
	\caption{
		An example for comparing multiple extracted SMs using the proposed evaluation metrics. The SM (B) has more Purity than the SM (A). The SM (C) has more Richness than SM (B) while having equal Purity. The SM (D) has less Scale and more Richness than SM (C).
	}
	\label{fig:sm_evaluation_metrics}
\end{figure}

The Scale metric measures the ratio of the number of SM's states that include final states to the total number of labels. A bigger Scale score means more sparsity of the space. Based on Equation~\ref{scale_eq}, the best Scale occurs when the number of SM's states, including PM's final states, equals the number of class labels. As illustrated in Fig.~\ref{fig:sm_evaluation_metrics}, the SM (D) has a better scale than the SM (C) because of having two SM states, including the PM's final states. A Scale score lower than one means a lack of discrimination between the PM's final states with different labels, which is unacceptable.

\begin{equation}\label{scale_eq}
Scale = \frac{\sum_{\|SMS\|}^{i=1} \| \{i\;|\;s \in SMS[i], s.pl \neq Null \} \|}{\|L\|}
\end{equation}

As an example, let's consider the states machines SM (C) and SM (D) in Fig.~\ref{fig:sm_evaluation_metrics}. For SM (C), there are three states (S2, S5, and S6) that include PM's final states, whereas for SM (D), only two states (S4 and S6) include final states.

Assuming there are two class labels (blue and green) in the primary model, the Scale scores for the state machines would be calculated as follows:

For SM (C): 
\begin{equation*}
Scale = \frac{3}{2}=1.5
\end{equation*}

For SM (D):
\begin{equation*}
Scale = \frac{2}{2}=1
\end{equation*}

Therefore, based on the Scale metric, SM (D) is simpler and more discriminative and thus has a better scale than SM (C). This is because it has fewer states that include the PM's final states, but still successfully distinguishes between the different class labels of the PM's final states. On the other hand, a Scale score larger than one, as in the case of SM (C), indicates unnecessary complexity in the model and a sparse representation of the state space.

\subsection{Coverage Criteria}
There are different approaches for generating and evaluating test suites for neural networks~\cite{huang2020survey}. As mentioned in Section~\ref{sec:related}, there are two categories of approaches. One group of works propose different mutation operators and evaluate a test suite by measuring the rate of killed mutants~\cite{shen2018munn, cheng2018manifesting, ma2018deepmutation}. In the other category, different coverage criteria in a specific context are defined and the coverage is measured in specified areas~\cite{tian2018deeptest, pei2017deepxplore, sun2018concolic, wicker2018feature, kim2019guiding, ma2018deepgauge, du2019deepstellar}. Utilizing most of the mentioned coverage criteria in practice is challenging due to the inherent complexity of neural networks.  In this paper, we take advantage of the extracted SM to define new coverage criteria in a simpler model. The proposed coverage criteria measure the coverage of a test suite on SM's specified states. 

\subsubsection{New Final State Coverage}
As mentioned in Section~\ref{sec:extraction}, a subset of the SMS (states of the derived state mechine) include at least one final state of the primary model. These states are considered as the final states of the state machine. If an input is terminated on a non-final state of the state machine, it could potentially be an input that exposes unforeseen behavior in the primary neural network.

Our first coverage criteria, New Final State Coverage (NewFSCov), measures the percentage of the non-final states of the SM which include a terminating state by at least one test data. More formally, we define SMFS and SMFS' as the set of states in the state machine including at least one final state from the training data and test data respectively (Equations~\ref{SMFS_definition_eq} and~\ref{SMFS'_definition_eq}). NewFSCov can then be defined as Equation~\ref{NewFSCov_definition_eq}.

\begin{equation}\label{SMFS_definition_eq}
    SMFS = \{ S \; | \;  S \in SMS, \exists fs \in S : fs \in PMFS  \}
\end{equation}
\begin{equation}\label{SMFS'_definition_eq}
    SMFS' = \{ S \; | \; S \in SMS, \exists fs \in S :  fs \in PMFS'  \}
\end{equation}
\begin{equation}\label{NewFSCov_definition_eq}
	NewFSCov = \frac{\| SMFS' - SMFS \|}{\| SMS - SMFS \ \|} 
\end{equation}

\subsubsection{Out of Boundary Final State Coverage}
As mentioned in Section~\ref{sec:extraction}, using an extracted state machine for a test data may lead to new state generation for the state machine. In other words, out of boundary states are those that are not defined during the SM extraction phase on the training data. A test data which terminates in a new generated state can be a candidate for revealing unexpected behaviour of the primary model.
The Out of Boundary Final State Coverage (OutFSCov) metric attempts to evaluate a test suite from this perspective. As defined in Equation~\ref{OutFSCov_eq}, this metric calculates the number of new final states discovered by the test suite.

\begin{equation}\label{OutFSCov_eq}
OutFSCov = \| SMFS' - SMS \|
\end{equation}

\subsubsection{Basic Final State Coverage}
The term ``Basic'' refers to the states identified during the SM extraction process on the training data. The Basic Final State Coverage (BasicFSCov) metric is used to measure the coverage of the  basic states by the test suite.

\begin{equation}
BasicFSCov = \frac{\| SMFS' \cap SMFS \|}{\| SMFS \|} 
\end{equation}

\subsubsection{Basic Label and Final State Coverage}
The next criteria to evaluate the test suite is to measure the percentage of the SM final states covered by the test data. In this measurement, we also incorporate the predicted class labels to ensure that the test suite will visit the final states that have similar predicted labels to the training data.
More formally, we define pairs of labels and final states generated by the training data and test data on the primary model as in Equations~\ref{PMLFS_eq} and~\ref{PMLFS'_eq}, respectively. Accordingly, the corresponding states and labels in the SM are defined using on $PMLFS$ and $PMLFS'$ as in Equations~\ref{SMLFS_eq} and~\ref{SMLFS'_eq}. As previously mentioned, SM final states are those which include at least one PM final states. Basic Label and Final State Coverage (BasicLFSCov) attempts to measure the coverage of the test suite on basic final states with the same predicted class labels, as expressed in Equation~\ref{BasicLFSCov}. 

\begin{equation} \label{PMLFS_eq}
PMLFS = \{(fs.pl,fs) \; | \;  fs \in PMFS  \} 
\end{equation}

\begin{equation} \label{PMLFS'_eq}
PMLFS' = \{(fs.pl,fs) \; | \;  fs \in PMFS'  \} 
\end{equation}

\begin{equation} \label{SMLFS_eq}
SMLFS = \{(l,S) \; | \; \exists s : (l,s) \in PMLFS, s \in S   \} 
\end{equation}

\begin{equation} \label{SMLFS'_eq}
SMLFS' = \{(l,S) \; | \;  \exists s : (l,s) \in PMLFS', s \in S \} 
\end{equation}

\begin{equation} \label{BasicLFSCov}
BasicLFSCov = \frac{\| SMLFS \cap SMLFS' \| }{\| SMLFS \|}
\end{equation}

\subsubsection{Weighted Basic Label and Final State Coverage}
The existing metric can be adjusted by incorporating the number of states within each state machine (SM) state. Our objective is to assign higher significance to label-state pairs that were more frequently observed in the training data.
More formally, the weight function, $W(l,S)$ returns the number of training data points in each SM state for each predicted class label (Equation~\ref{weight_function_eq}). By incorporating this weight function into coverage measurement, WeightedBasicLFSCov is defined as in Equation~\ref{WeightedBasicLFSCov_eq}.

\begin{equation} \label{weight_function_eq}
W(l,S) = \| \{ s \; | \; s \in S, (l,s) \in PMLFS\} \|
\end{equation}

\begin{equation} \label{WeightedBasicLFSCov_eq}
WeightedBasicLFSCov = \frac{\sum_{(l,s) \in  (SMLFS \cap SMLFS') } W(l,s) }{\sum_{(l,s) \in  SMLFS}W(l,s)}
\end{equation}

\subsubsection{Weighted Label and Final State Coverage}
The next metric measures the coverage of SM states and labels with increased emphasis on those  that have not been visited or visited less frequently by the training data. More formally,  the weight function $W'(l,s)$ is defined to assign higher weights to those pairs of label and state that are visited less frequently (Equation~\ref{inverse_weight_function_eq}). 
The Weighted Label and Final State Coverage (WeightedLFSCov) metric is then established as per Equation~\ref{WeightedLFSCov_eq}, to aggregate the weights of all pairs visited by the test data.

\begin{equation}\label{inverse_weight_function_eq}
W'(l,S) = \frac{1}{W(l,S)+1}
\end{equation}

\begin{equation} \label{WeightedLFSCov_eq}
WeightedLFSCov = \sum_{(l,S) \in SMLFS'} W'(l,S)
\end{equation}

The WeightedLFSCov and OutFSCov metrics are not proportional due to the potential for an unbounded number of SM states, as new marginal states may be created by test data.

\subsection{Online Error Prediction}
\label{online_error_pred_sec}
By online error prediction, we refer to predicting the error probability of the primary model for each input in real-time, as new input data is processed. This helps to detect potential errors in the primary model before they occur, allowing for more precise predictions and improving the model performance. By performance, we mean the PM's accuracy in various situations in terms of internal state and state transitions. Utilizing the extracted state machine, we propose an approach for online error prediction of an RNN-based model. The idea is to train a tree-based model based on features extracted from the state space of the extracted SM. The labels utilized to train the tree-based model are derived from errors made by the primary model on a test dataset. The advantage of using a tree-based model is to use its expressive power to have an understanding of the reason behind a predicted error. In other words, using the rules in the tree, we can explain why the tree has marked an input as a candidate for an error.

A tree-based model predicts the value of a target variable by learning simple decision rules inferred from the data features. The model can be considered as a piece-wise constant approximation. Decision tree is a non-parametric supervised learning method for creating a tree-based model~\cite{loh2011classification}. The decision tree method employs various algorithms to identify the optimal feature to split and generate the rules. One widely used algorithm is Iterative Dichotomiser 3 (ID3), proposed by Ross Quinlan in 1986~\cite{quinlan1986induction}. 
In a greedy fashion, ID3 constructs a multiway tree by identifying the feature  that maximizes information gain and minimizes entropy at each node.
 Trees are typically grown to their maximum size and then pruned to improve their ability to generalize to unseen data.

We train a tree-based model based on the faulty behaviour of the primary model to predict the errors at service time.  Labels for incorrect outputs of the primary model are assigned a value of zero, while correct outputs are assigned a value of one. 
The features for training the tree-based model are defined using the extracted state machine.  When the state machine is run on an input $i$, a sequence of states are traversed through a sequence of transitions. We define $States(i)$ and $Trans(i)$ as functions that return the set of states and transitions traversed by the input $i$ on the state machine. $fState(i)$ returns the final state of the state machine visited by $i$. The predicted label of the input $i$ by the primary model is represented by $pLabel(i)$. In the following, $SMS$ and $SMT$ refer to the set of states and transitions of the state machine (extracted from the training data), respectively. The details of the extracted features  are as follows:

\begin{enumerate}
    \item New Transitions  \\
    This feature assesses the presence of transitions in the trace of states for a given input ($i$) that are absent from the transitions in the training dataset. If the trace contains a transition from one state to another that has not been previously observed in the training dataset, the feature will have a value of one. For each additional previously unobserved transition, the value will increment accordingly. If no such transitions are present, the feature will have a value of zero. Considering $NT$ to be a function returning the value of ``new transitions'' for an input $i$, we have the following:
    
    \begin{equation}
    NT(i) = \| Trans(i) - SMT  \|
    \end{equation}
    
    \noindent As mentioned, $SMT$ is the set of transitions of the state machine extracted from the training dataset. The above equation calculates the size of the set of transitions traversed by the input $i$ that have not been already present in the state machine.

    \item New States  \\
     Analogous to the aforementioned characteristic, this feature measures the number of  newly encountered states traversed by the input.
     More formally, the ``new states'' feature can be defined as a function $NS$ on an input $i$ as follows:
    
    \begin{equation}
    NS(i) = \| States(i) - SMS  \|
    \end{equation}

    \item Final State Share Rate  \\
    Each state of the state machine may include different final states of the primary model when run by the training data with different predicted labels. The next feature is to evaluate the final state machine state of an input ($fState(i)$) by the percentage of the primary model final states in that state with similar predicted label. The less percentage of similarity, the more likelihood of unexpected prediction by the primary model. 
     More formally, the ``Final State Share Rate'' feature can be defined as a function $FSSR$ on an input $i$ as follows: 
    \begin{equation}
    FSSR(i) = \frac{\sum_{s \in fState(i)}^{} (s.pl = pLabel(i))  }{\sum_{s \in fState(i)}^{} (s.pl \neq Null ) }
    \end{equation}
    
    \noindent, where the $pl$ is the predicted label by the PM for each state. $pLabel$ returns the label of the input $i$ predicted by the primary model.

    \item Count at Final State ($CFS$) \\
    The previous feature gave an insight into the proportional rate of the final states with similar label. However, if the number of final states in a state machine state is substantially low, then a high proportional rate can be misleading. This is primarily because a limited number of final states suggests that the state machine state has not been adequately visited.  Consequently, we employ an absolute measure to offer a more comprehensive insight in conjunction with the preceding feature. More formally, the ``Count at Final State'' feature can be defined as a function $CFS$ on an input $i$ as follows: 
    \begin{equation}
    CFS(i) = \sum_{s \in fState(i)}^{} (s.pl = pLabel(i)) 
    \end{equation}
    
    \item Trace Probability ($TP$) \\
    The next feature measures the probability of the trace of transitions traversed by the input based on the training data.  Intuitively, an input sequence with a lower probability suggests a higher chance of the primary model producing an error in the output. To calculate the trace probability, we assign a probability to each transition of the state machine based on the training data. Defining  a transition $t$ by its source and target states in the state machine as $(S_i \rightarrow S_j)$, the probability of $t$, represented by $P(t)$ can be calculated as the ratio of the number of transitions from $S_i$ to $S_j$ to the overall count of outgoing transitions from $S_i$. These counts are determined based on the training data behavior. 
     Having the transition probability, the probability of the sequence of transitions traversed by the input $i$ can be defined as the multiplication of the probabilities of all transitions in the sequence (Equation~\ref{tp_eq})
     
    \begin{equation}\label{tp_eq}
    TP(i) = \prod_{t \in Trans(i)} P(t)
    \end{equation}
    
    \item Transitions Probability Mean ($TPM$) \\
    If an input $i$ traverses a novel transition in the state machine, then the  probability of that transition equals zero. Based on Equation~\ref{tp_eq}, the trace probability will be zero, regardless of the probability of other transitions in the trace. To take probability of other transitions into account in these situations, we define a new feature to measure the mean of the transitions probabilities in the sequence.   This feature allows for a deeper understanding of the characteristics and behavior of novel traces by considering not only the absence of prior knowledge, but also the distribution of transition probabilities within the trace. The $TPM$ is defined as follows:
    \begin{equation}
    TPM(i) = \frac{1}{\| Trans(i) \|} \sum_{t \in Trans(i)}^{} P(t)
    \end{equation}

\end{enumerate}

The analysis of the computational overhead associated with the training of the tree-based classifier for online error prediction is as follows. The training data consists of input sequence features and error labels extracted from state machine traces over the validation dataset. Gathering these training instances takes $O(\|TD\| * MTS)$ time where $\|TD\|$ is dataset size and $MTS$ is sequence length. Decision tree induction algorithms like ID3 have a time complexity of $O(\|TD\| * F^2)$ for training, where $F$ is the number of features. This is because at each tree node, the algorithm checks every feature value against every data point to determine the optimal split \cite{loh2011classification}. However, given that the number of features is not excessively large, the time complexity can be considered linear.

\section{Experimental Results}
\label{sec:experiments}
In this section, we present the results of our experiments for evaluating our proposed methodologies: 1) extracting SM from RNN-based models, 2) coverage criteria based on the extracted SM for evaluating generated test suites, and 3) utilizing the extracted SM to predict potential errors in the PM. Through conducting experiments, we aim to gain a deeper understanding of the quality of the proposed methodologies. The datasets used in our experiments include MNIST \cite{lecun1998gradient} and Mini Speech Commands \cite{warden2018speech}, which are utilized to train RNN-based models. The RNN-based models employed in the study utilize different recurrent modules, including GRU, LSTM, and S-RNN.

MNIST is a benchmark dataset in the field of computer vision and machine learning, used extensively for the evaluation of image recognition algorithms. The dataset consists of 70,000 training images of handwritten digits (0-9), each represented as a 28$\times$28 grayscale image. We chose the MNIST dataset because it provides a simpler test case for evaluating our SM extraction and coverage testing techniques. The images consist of single, centered digits, allowing us to train models with reasonable accuracy. This enables analysis to focus more on assessing the effectiveness of the proposed internal state-based testing approach rather than overcoming challenges of highly complex models or datasets. The Mini Speech Commands dataset is a widely used benchmark for speech recognition systems. It comprises 65,000 one-second-long utterances of eight simple English words, including "yes", "no", "up", "down", "right", "left", "stop", and "go", spoken by a diverse group of individuals. Commands like these play a significant role in applications such as voice-controlled electronic wheelchairs or home automation systems. However, incorrect identification of these vital voice commands can lead to severe implications for the users. Therefore, it is crucial that we thoroughly analyze and test the logic of speech recognition AI models like the ones trained on this dataset and uncover potential errors during the service time. A summary of the details for the training datasets is presented in Table. \ref{datasets_details_tbl}.

\begin{table}[ht]
\caption{Details of our datasets}\label{datasets_details_tbl}
\begin{tabular*}{0.65\textwidth}{@{}lll@{}}
\toprule
Detail & MNIST & Mini Speech Commands \\
\midrule
Number of examples & 70,000 images & 65,000 utterances \\
Input dimensions & 28 x 28 grayscale images & 124 x 129 spectrograms \\
Number of classes & 10 digits & 8 commands \\
\bottomrule
\end{tabular*}
\end{table}

The RNN-based models for MNIST and Mini Speech Commands datasets have been trained with the following layer configurations: $[(28,28),64,10]$ and $[(124,129),64,8]$, respectively. The first component of the layer configurations, $(28,28)$ and $(124,129)$, represents the input layer dimensions. The MNIST dataset consists of 28$\times$28 grayscale images, while the Mini Speech Commands dataset consists of 1-second audio recordings that have been transformed from the time domain to the frequency domain, resulting in 124$\times$129 size representations. The second component, 64, denotes the size of the recurrent layer. In the case of multiple recurrent layers, the state vectors are concatenated horizontally at each time step. The final component, 10 and 8, respectively, represents the number of classes in the output layer. The models were trained using the RMS-Prop optimization algorithm \cite{tieleman2012lecture} with categorical cross entropy loss. The learning rate was initially set to 0.1 and reduced after some epochs down to 0.0001. A batch size of 32 was used for training all models. The number of epochs was around 20 for the GRU and LSTM models, and 60 for the S-RNN model. The models were initialized with random weights rather than pre-trained weights. Training was performed using an NVIDIA Tesla T4 GPU with 16GB of GDDR6 memory and 2560 CUDA cores, coupled with 12.7GB of RAM. The training process for each model took less than an hour.

The trained models exhibit an accuracy greater than 95\% on both the training and test datasets. However, the model utilizing the S-RNN recurrent module has demonstrated poor accuracy on the Mini Speech Commands dataset, and as a result, has been disregarded in experiments related to the Mini Speech Commands dataset.

\subsection{State Machine Extraction and Evaluation}
In this section, we present the experimental results of evaluating our method for state machine extraction. We then compare the results with those obtained by the DeepStellar method~\cite{du2019deepstellar}. We use our proposed metrics for the state machine evaluation purpose. 

As there is no single mathematical definition of explainability, Molnar provides a useful non-mathematical concept: the degree to which a human can understand the cause of a model's decisions \cite{molnar2018interprtable}. Explainability can be evaluated via human studies on simple tasks, or by proxy using models like state machines that are considered interpretable. Our state machine extraction leverages the latter, providing explainability without direct human evaluation. The understandability of state machines allows us to better comprehend the primary model’s logic. Further, as seen in the subsequent subsections, this explainability enabled proposing meaningful coverage criteria that statistically significantly target important model behaviors. It also facilitated extracting insightful features for real-time error prediction - an otherwise opaque task. In this way, the state machine explainability paved the path for enhanced testing and monitoring of the black-box recurrent neural network.

DeepStellar is a well-established approach for extracting state machines from deep learning models. By comparing the performance of our proposed method with DeepStellar, we aim to demonstrate the advantages and improvements offered by our approach in terms of the quality of the extracted state machines and their usefulness for evaluating test suites and predicting errors in the primary model.
The proposed SM extraction method differs from DeepStellar's method in the way it defines the states of the SM. Our method uses K-Means clustering algorithim which considers the patterns of the data and defines states based on the distribution of PM states in the state space. In contrast, DeepStellar defines states using a grid-based approach, dividing the state space into hyper-cubes that are considered as SM states. However, this static definition of SM states can lead to an enormous number of states in high-dimensional state spaces. To address this issue, DeepStellar employs the Principal Component Analysis (PCA) algorithm to reduce the number of dimensions in the state space.
PCA~\cite{jolliffe2002principal} is a statistical technique that is commonly used for dimensionality reduction of multivariate data. The goal of PCA is to reduce the dimensionality of the data while retaining as much of the variation in the data as possible. In other words, PCA tries to identify a set of orthogonal, linearly uncorrelated variables known as principal components, that can capture the most important patterns and structures in the data.

In our experiments, we also employ Linear Discriminant Analysis (LDA) for dimension reduction. This is done to facilitate a more effective comparison between an improved version of the DeepStellar method and our proposed method.LDA \cite{fisher1936use} is a dimensionality reduction and classification technique that finds the linear combination of features that maximizes the separation between classes. The main difference between PCA and LDA is that PCA is an unsupervised technique for dimensionality reduction, while LDA is a supervised technique that also considers class discrimination. PCA focuses on finding the directions of maximum variance in the data, regardless of the class labels, while LDA finds the linear combination of features that maximizes the separation between classes. We train an LDA model on the final PM states with their predicted class labels. Subsequently, this LDA model is used to reduce the dimensions of the non-final states as well. This technique helps to decrease the state space dimensions with respect to the  class labels predicted by the primary RNN-based model.

In the process of extracting SMs, the optimal hyperparameters for the SM extraction algorithms are employed. These optimal hyperparameters are determined through the utilization of the proposed SM evaluation metrics. This enables us to compare the best probable SMs extracted from each model, using the aforementioned SM extraction algorithms. 

Throughout the rest of this paper, we will refer to our proposed method as "DeepCover". The original state machine extraction method of DeepStellar, which uses grid-based clustering and PCA dimension reduction, will be  simply referred to  as "DeepStellar". The improved version of DeepStellar, where PCA is replaced with LDA, will be referred to as "DeepStellar-LDA". 

To extract the SM, the trained RNN models are fed with the full training datasets (MNIST and Mini Speech Commands). The internal state vectors of the RNN models are collected at each timestep. These state vectors represent points in the RNN model's state space. K-Means clustering is then applied on these state vectors to cluster them into a discrete set of states for the extracted state machine. For DeepStellar's approach, a grid is imposed on the state space and each grid cell is considered a state. To reduce the dimensionality of the state space, PCA is used by DeepStellar while our proposed DeepCover approach uses LDA, which considers class labels during dimensionality reduction. Overall, the training data traces are used to extract a representative state machine that mimics the RNN model's behavior. Tables~\ref{sm_extraction_comparison_image_tbl} and~\ref{sm_extraction_comparison_speech_tbl} present a comparative study of the state machines extracted from primary RNN-based models trained on the MNIST and Mini Speech Commands datasets, respectively, using different recurrent modules. The tables evaluate the performance of the state machine extraction methods using the proposed metrics: Purity, $Richness$, $Goodness$, and $Scale$. In addition, the number of SM states presented in Tables~\ref{sm_extraction_comparison_image_tbl} and~\ref{sm_extraction_comparison_speech_tbl} indicates the number of hyper-cubes in the grid-based method for the area where the RNN state vectors are located. It also shows the number of clusters in the K-Means method used for discretizing the state space of the RNNs.

In extracting an explainable model from a primary model, a core objective is for the secondary model to mimic the behavior and functionality of the primary one as closely as possible. The proposed $Purity$, $Richness$, and $Scale$ metrics aim to quantitatively validate how well these complex patterns learned by the primary model are preserved. Specifically, $Purity$ checks that the separation of final state predicted labels by primary model is maintained, ensuring states discriminate between classes as intended. $Richness$ verifies that final states are adequately grouped into relevant states based on their predicted labels, avoiding unnecessary fragmentation. Additionally, $Scale$ confirms states adequately encompass the range of predicted labels without over-engineering complexity, supporting interpretability. A $Scale$ score lower than one indicates a suboptimal accuracy of the SM in classification task. As previously described, the $Goodness$ metric is mathematically expressed as $Goodness = Purity^{10} * Richness$. It can be deduced that if certain extracted SMs possess low values of $Purity$, but high values of $Richness$, the resulting $Goodness$ score will be relatively low. By extracting state machines that exhibit high scores on these metrics, we can gain a deeper understanding of the complex, often opaque, inner workings of the recurrent neural network.

The results indicate that DeepCover consistently produces high-quality state machines, as shown by the $Goodness$ scores. For instance, in Table~\ref{sm_extraction_comparison_image_tbl}, the $Goodness$ score of state machine extracted from the LSTM primary model using DeepCover is 1236.7, significantly higher than when using the DeepStellar and DeepStellar-LDA. Similarly, in Table \ref{sm_extraction_comparison_speech_tbl}, the $Goodness$ score of the GRU primary model using DeepCover is 337, again outperforming the DeepStellar.
The $Scale$ score also shows the DeepCover's ability to maintain a reasonable state machine size, which is crucial for interpretability. The results show that DeepCover ensures a balance between the complexity of the model (as indicated by the $Scale$ score) and its performance (as indicated by the $Goodness$ score), demonstrating its effectiveness in enhancing the explainability of RNN-based primary models.
The tables also highlight the impact of the dimension reduction technique on the quality of the extracted SM. The usage of LDA in DeepStellar-LDA for some cases has led to better results compared to DeepStellar, suggesting that considering class labels in dimension reduction can enhance the quality of the derived state machine.

Lastly, observing the number of states in the state machines extracted by DeepCover provides further insight. For the GRU and LSTM models, DeepCover extracted state machines with only 25 states, while for the S-RNN models, 100 states were extracted. Considering that these models operate in a continuous 64-dimensional state space (RNN module with 64 dimensional information processing band), reducing them to interpretable state machines with 25-100 discrete states demonstrates DeepCover's ability to greatly simplify complex recurrent neural networks while maintaining their key functionality. The conciseness of the extracted state machines enables enhanced explainability by providing a clear yet representative abstraction of the primary model's internal decision-making process. The reduced number of states also allows for more tractable analysis and testing of the model through the definition of coverage criteria and error prediction mechanisms using the simplified state machine representation. Overall, DeepCover's state machine extraction strikes an effective balance between model performance and interpretability through controlled state space discretization.

\begin{table}[ht]
\caption{Comparison of the methods for state machines extraction from RNN-based primary models trained on the MNIST dataset}\label{sm_extraction_comparison_image_tbl}
\begin{tabular*}{0.8\textwidth}{@{}lllllll@{}}
\toprule
SM Extraction Method & RNN Type & Purity & Richness & Goodness & Scale & \begin{tabular}{@{}c@{}}SM States \\ on TD\end{tabular} \\ 
\midrule
DeepStellar & LSTM & 76 & 163 & 10.7 & 36.8 & 655 \\
DeepStellar-LDA & LSTM & 90 & 540 & 183.6 & 11 & 496 \\
DeepCover & LSTM & 93 & 2500 & 1236.7 & 2.4 & 25 \\
\hline
DeepStellar & S-RNN & 49 & 4285 & 3.4 & 1.4 & 21 \\
DeepStellar-LDA & S-RNN & 81 & 1225 & 147.6 & 4.9 & 54 \\
DeepCover & S-RNN & 94 & 833 & 489 & 7.2 & 100 \\
\hline
DeepStellar & GRU & 19 & 15000 & 0 & 0.4 & 7 \\
DeepStellar-LDA & GRU & 78 & 2143 & 153 & 2.8 & 61 \\
DeepCover & GRU & 97 & 3000 & 2267.6 & 2 & 25 \\
\bottomrule
\end{tabular*}
\end{table}

\begin{table}[ht]
\caption{Comparison of the methods for state machines extraction from RNN-based primary models trained on the Mini Speech Commands dataset}\label{sm_extraction_comparison_speech_tbl}
\begin{tabular*}{0.8\textwidth}{@{}lllllll@{}}
\toprule
SM Extraction Method & RNN Type & Purity & Richness & Goodness & Scale & \begin{tabular}{@{}c@{}}SM States \\ on TD\end{tabular} \\ 
\midrule
DeepStellar & LSTM & 99 & 1.4 & 1.2 & 561 & 187660 \\
DeepStellar-LDA & LSTM & 89 & 52 & 16.5 & 16 & 521 \\
DeepCover & LSTM & 93 & 256 & 129 & 3.1 & 25 \\
\hline
DeepStellar & GRU & 36 & 1280 & 0.05 & 0.6 & 14 \\
DeepStellar-LDA & GRU & 99 & 31 & 26.8 & 26 & 1324 \\
DeepCover & GRU & 98 & 427 & 337 & 2 & 15 \\
\bottomrule
\end{tabular*}
\end{table}

\subsection{Coverage Criteria}
In Section~\ref{sec:methodology}, we introduced the concept of coverage criteria as a method for evaluating the effectiveness of test suites in uncovering defects in the primary model. These criteria provide a set of metrics to measure the completeness of testing with respect to the SM extracted from the primary model. In this section, we aim to validate the efficacy of these coverage criteria by performing a statistical test. The purpose of this test is to assess the significance of the coverage criteria and determine their ability to accurately measure the quality of generated test suites.

We have used Kolmogorov-Smirnov test~\cite{kolmogorov1933sulla} to gain a deeper understanding of the relationship between the coverage criteria and the effectiveness of generated test suites in uncovering defects. The Kolmogorov-Smirnov test compares the cumulative distribution functions (CDFs) of two distributions by calculating the maximum distance between them, which we represent by $D$.  The p-value represents the probability that the difference of CDF as extreme as $D$ could have occurred by chance, under the null hypothesis that the two samples were drawn from the same distribution.  Typically, a p-value less than 5\% indicates a statistically significant difference between the cumulative distribution functions of two distributions. In the following, we describe the test method and its results.

A statistical test has been designed, drawing upon the methodology of the Kolmogorov-Smirnov test. For each proposed coverage criterion, 1000 test suites, each containing 1000 test inputs for the primary model, have been generated. Test case generation has been performed on the MNIST and Mini Speech Commands datasets using random transformations such as rotation, zoom-in, and zoom-out for MNIST, and speeding, pitching, and adding noise for the Mini Speech Commands. The state vectors for each test input at each time step have been extracted from the primary model, and the resulting traces on the SM have been obtained. Using this information, the value of coverage criteria for each test suite is calculated.

\begin{figure}[ht]
	\centerline{\includegraphics[width=0.7\textwidth]{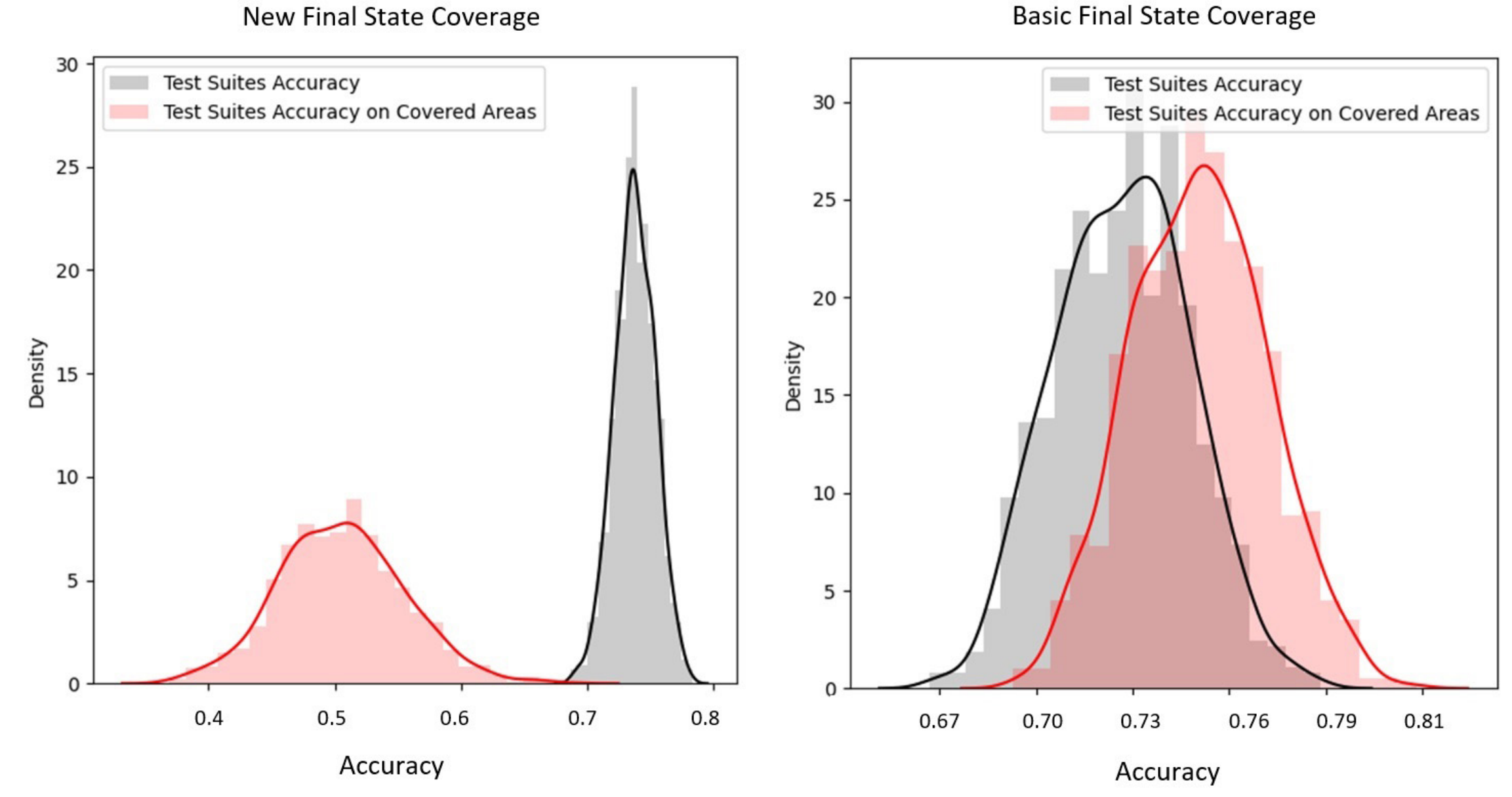}}
	\caption{
        Comparison of accuracy distributions for targeted and all areas. 
    }
	\label{fig:test_suits_accuracy}
\end{figure}

\iffalse
The proposed coverage criteria aim to classify targets into two types: generalization targets and fidelity targets. Generalization targets are difficult for the model to distinguish new patterns, while fidelity targets are easy because they have been trained on the training data. 
\fi
The significance of each coverage criterion is evaluated by comparing the accuracy distribution of the test suites with the accuracy distribution of the test suite subsets that cover the areas defined by each criterion. For instance, in the case of $New Final State Coverage$, the accuracy distribution of the test suites is compared with the accuracy distribution of the test suite subsets that reach new final states.
If the accuracy measures of the test suites on the covered areas show a substantial deviation from the accuracy measures on all areas, then we can claim that the coverage criterion has been a good measure for the quality of the test suites. 
As an example, Fig.~\ref{fig:test_suits_accuracy} visualizes the difference in accuracy distributions between targeted areas and all areas for $NewFinalStateCov$ and $BasicFinalStateCov$ coverage criteria. The red curves represent the distribution of accuracy on targeted areas, while the grey curve represents the distribution of accuracy on all areas. The Kolmogorov-Smirnov test was used to assess the similarity of these distributions, with the null hypothesis being that they are drawn from the same underlying distribution. 

We have also evaluated  the coverage criteria proposed by DeepStellar~\cite{du2019deepstellar} by running the Kolmogorov-Smirnov test. The coverage criteria proposed by DeepStellar are as follow:
\begin{enumerate}
    \item 
    $BasicSCov$, which measures the coverage of the state machine states. An SM state is considered to be covered if it is visited by the primary model states on both the test suite and the training data.
    \item 
    $WeightedSCov$, which extends $BasicSCov$ by adding weights to each SM state based on the frequency of visits by the primary model states on training data, thereby providing more importance to less frequently visited states.
    \item 
    $OutSCov$, which considers an SM state to be covered if it is visited by the primary model states, while being outside of the state space explored during training. This criterion ensures that the primary model is capable of handling inputs that fall outside of its trained state space.
    \item 
    $BasicTCov$, which defines a coverage metric as a binary criterion that considers an SM states transition covered if it is taken by the primary model on both training data and test suite.
    \item 
    $WeightedTCov$, which extends $BasicTCov$ by adding weights to each transition based on the frequency of the primary model taking the transition on training data, thereby providing more importance to less frequently taken transitions.
\end{enumerate}

The outcomes of the Kolmogorov-Smirnov tests are reported in tables~\ref{cov_test_image_tbl} and~\ref{cov_test_speech_tbl} for the MNIST and Mini Speech Commands datasets, respectively. The symbol \xmark\ indicates a p-value greater than 0.05, implying no substantial disparity in the accuracy distribution of test suites between all areas and targeted areas. The symbol \cmark\ , on the other hand, indicates a significant difference, and hence, the efficacy of the proposed coverage criteria. 

\begin{table}[ht]
\caption{Results of the Kolmogorov-Smirnov test on different coverage criteria conducted on the MNIST dataset}\label{cov_test_image_tbl}
\begin{tabular*}{0.8\textwidth}{@{}lccccccc@{}}
\toprule
 & Coverage & \multicolumn{3}{c}{Extracted by DeepStellar} & \multicolumn{3}{c}{Extracted by DeepCover} \\
 &  Criteria by & S-RNN & LSTM & GRU & S-RNN & LSTM & GRU \\ 
\midrule
$NewFSCov$ & DeepCover & \cmark & \cmark & \cmark & \cmark & \cmark & \cmark \\
$OutFSCov$ & DeepCover & \cmark & \cmark & \cmark & \cmark & \cmark & \cmark \\
$BasicFSCov$ & DeepCover & \xmark & \cmark & \xmark & \cmark & \xmark & \cmark \\
$BasicLFSCov$ & DeepCover & \cmark & \cmark & \cmark & \cmark & \cmark & \cmark \\
$WeightedBasicLFSCov$ & DeepCover & \cmark & \cmark & \cmark & \cmark & \cmark & \cmark \\
$WeightedLFSCov$ & DeepCover & \cmark & \xmark & \xmark & \xmark & \xmark & \cmark \\
$BasicSCov$ & DeepStellar & \xmark & \cmark & \xmark & \xmark & \xmark & \xmark \\
$WeightedSCov$ & DeepStellar & \cmark & \xmark & \xmark & \xmark & \xmark & \xmark \\
$OutSCov$ & DeepStellar & \cmark & \cmark & \cmark & \cmark & \cmark & \cmark \\
$BasicTCov$ & DeepStellar & \cmark & \cmark & \cmark & \cmark & \cmark & \cmark \\
$WeightedTCov$ & DeepStellar & \cmark & \cmark & \cmark & \cmark & \cmark & \cmark \\
\bottomrule
\end{tabular*}
\end{table}

\begin{table}[ht]
\caption{Results of the Kolmogorov-Smirnov test on different coverage criteria conducted on the Mini Speech Commands dataset}\label{cov_test_speech_tbl}
\begin{tabular*}{0.78\textwidth}{@{}lccccc@{}}
\toprule
 & Coverage & \multicolumn{2}{c}{Extracted by DeepStellar} & \multicolumn{2}{c}{Extracted by DeepCover} \\
 & Criteria by & LSTM & GRU & LSTM & GRU \\ 
\midrule
$NewFSCov$ & DeepCover & \cmark & \cmark & \cmark & \cmark \\
$OutFSCov$ & DeepCover & \cmark & \xmark & \cmark & \cmark \\
$BasicFSCov$ & DeepCover & \cmark & \cmark & \xmark & \cmark \\
$BasicLFSCov$ & DeepCover & \cmark & \cmark & \cmark & \cmark \\
$WeightedBasicLFSCov$ & DeepCover & \cmark & \cmark & \cmark & \cmark \\
$WeightedLFSCov$ & DeepCover & \xmark & \xmark & \cmark & \cmark \\
$BasicSCov$ & DeepStellar & \cmark & \cmark & \cmark & \cmark \\
$WeightedSCov$ & DeepStellar & \xmark & \xmark & \xmark & \cmark \\
$OutSCov$ & DeepStellar & \cmark & \xmark & \cmark & \cmark \\
$BasicTCov$ & DeepStellar & \cmark & \cmark & \cmark & \cmark \\
$WeightedTCov$ & DeepStellar & \cmark & \cmark & \cmark & \cmark \\
\bottomrule
\end{tabular*}
\end{table}

The results indicate that the coverage criteria proposed by DeepCover, namely $NewFSCov$, $BasicFSCov$, $BasicLFSCov$, $WeightedBasicLFSCov$, $BasicTCov$, and $WeightedTCov$, consistently demonstrate a significant difference in the accuracy distribution of test suites between all areas and targeted areas across both the MNIST and Mini Speech Commands datasets. This suggests that these criteria are effective in assessing the quality of test suites, regardless of the model or dataset used.
Here is an expanded explanation of what the significance means for each significant coverage criteria:
\begin{itemize}
    \item $NewFSCov$: The significance indicates reaching new final states not seen during training is importantly different than testing trained behavior. Covering unseen final states effectively targets model robustness - a crucial capability.
    \item $BasicFSCov$: Significance suggests re-covering key trained final state spaces is vital for validation, despite being well-explored. Focusing on these spaces that strongly influence predictions complements expanding testing to novel areas.
    \item $BasicLFSCov$: By considering state-label pairs, significance shows model accuracy depends heavily on whether learned label alignments in key state spaces hold on new inputs. Targeting these supports evaluating discrimination capability.
    \item $WeightedBasicLFSCov$: Incorporating state visitation frequencies, significance highlights that less frequent yet important learned label-state correlations must be re-confirmed during testing along with more common cases.
    \item $BasicTCov$: Transition coverage significance indicates thoroughly re-checking trained state transitions is imperative to effectively validate model logic, despite being repeated behaviors. Flaws here can undermine core functionality.
    \item $WeightedTCov$: Weighting transitions by rarity, significance means unusually infrequent transitions that may correspond to irregular control flows require explicit coverage too, not just prevalent cases. Targeting these can reveal corner case issues.
\end{itemize}

However, the criteria $OutFSCov$ and $WeightedLFSCov$ proposed by DeepCover, and $BasicSCov$, $WeightedSCov$, and $OutSCov$ proposed by DeepStellar, showed mixed results. This indicates that the effectiveness of these criteria may depend on the specific model and dataset used.

\subsection{Online Error Prediction}
In order to assess the efficacy of the online error prediction approach in identifying potential errors in the RNN-based model, we conducted an experiment utilizing input sequence datasets. We began by extracting the state machine from the RNN model and identifying features from the RNN model's state traces for each input data based on the extracted state machine and its state space, as detailed in Section~\ref{online_error_pred_sec}. Next, we trained a tree-based model using the extracted features and the errors made by the RNN model on the test dataset, with the goal of leveraging the tree-based model's ability to provide explanations for its error predictions. We employed the trained model to predict the error probability of the RNN model for each input sequence in the dataset.

For the purposes of our study, we trained the RNN-based model using the MNIST and Mini Speech Commands datasets. We obtained the labels for training the tree-based model by analyzing errors made by the RNN-based model on a generated test suite. To evaluate the accuracy of the tree-based model on a test suite, we utilized the area under the receiver operating characteristic curve (AUC).
The features extracted from the state space of the RNN-based model were designed to capture key aspects of the model's behavior, including the presence of previously unobserved transitions and states, the ability to differentiate between class labels in specific areas of the state space, and the predictability of input state sequences. Our objective was to identify potential errors in the RNN-based model during the service time through the training of a tree-based model on these features.

The results of our experiments, as presented in Table~\ref{auc_image_speech_tbl}, demonstrate the effectiveness of our online error prediction approach. Our method achieved high AUC values on the test datasets, indicating successful identification of potential errors. A key observation is that the features extracted from the state machines generated through our proposed DeepCover approach were more effective for the error prediction model compared to the state machine extraction method proposed by the DeepStellar study (i.e., griding the PM's state space). As shown in Table~\ref{auc_image_speech_tbl}, the use of DeepCover's state machine extraction resulted in higher AUC values for online error prediction across different recurrent modules and datasets.

\begin{figure}[ht]
	\centerline{\includegraphics[width=0.7\textwidth]{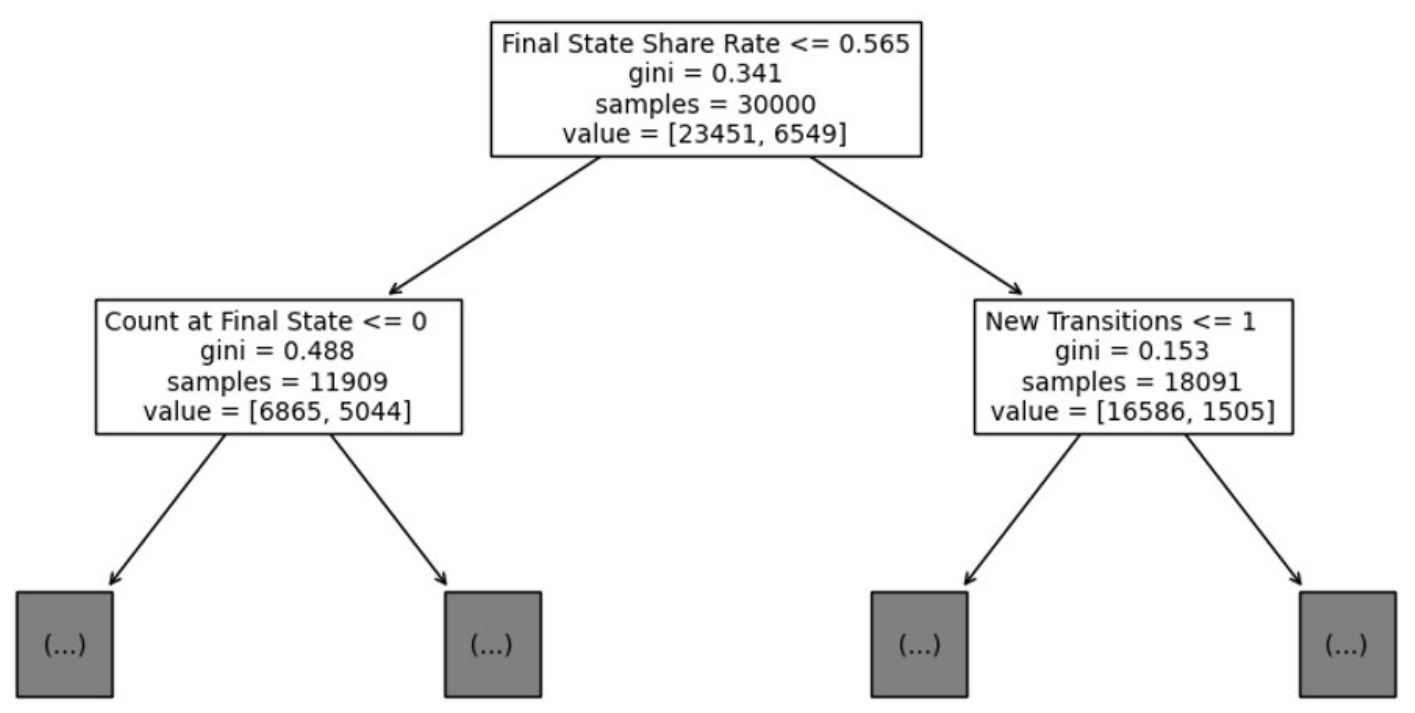}}
	\caption{The first-level of decision tree, derived from the trained tree-based model, illustrates the significant impact of certain features, namely Final State Share Rate, Count at Final State, and New Transitions, on decision-making processes. The rules on these features are also displayed.
    }
	\label{fig:tree_based_model_schema}
\end{figure}

Based on the Fig. \ref{fig:tree_based_model_schema}, the use of features extracted from the state space of the RNN-based model, in conjunction with the interpretability of the tree-based model, allowed us to obtain valuable insights into the behavior of the RNN-based model and accurately predict potential errors. Table~\ref{tab:feature-importance} presents the feature importance rankings of the extracted features, determined by their information gain with respect to error prediction. The $Final State Share Rate$ is the most important feature, with a relative importance score of 0.777. This indicates that the distribution of final states with matching versus differing predicted labels within the final state machine state visited by the input has the highest correlation with errors in the primary model. A lower share rate likely signals potential inaccurate predictions. The $Count at Final State$ has the next highest importance score of 0.134. This feature measures the absolute number of final states within the visited state machine state that have the same label as predicted for the input. A lower count suggests the visited state has not been explored adequately during training to develop a reliable prediction capability. The $New Transitions$ feature has a score of 0.050, suggesting transitions between state machine states that have been previously unobserved introduce moderate possibility of errors in the primary model's outputs. The $Transitions Probability Mean$ and $New States$ provide additional but more limited information to identify potential anomalies in the primary model's behavior with scores of 0.019 and 0.013 respectively. Finally, the $Trace Probability$ indicating likelihood of the overall input path through the state machine is the least correlated feature with a score of 0.006. Nevertheless, even small contributions from multiple features combine within the interpretable tree model to predict errors effectively.

\begin{table}[ht]
\caption{AUC results for different recurrent modules and state machine extraction methods}\label{auc_image_speech_tbl}
\begin{tabular*}{0.72\textwidth}{@{}llll@{}}
\toprule
Dataset & Recurrent Module & Clustering Method & \begin{tabular}{@{}c@{}}Online Error\\ Prediction AUC\end{tabular} \\ 
\midrule
MNIST & LSTM & DeepStellar & 0.83 \\
MNIST & S-RNN & DeepStellar & 0.79 \\
MNIST & GRU & DeepStellar & 0.8 \\
MNIST & LSTM & DeepCover & 0.92 \\
MNIST & S-RNN & DeepCover & 0.87 \\
MNIST & GRU & DeepCover & 0.9 \\
Mini Speech Commands & LSTM & DeepStellar & 0.85 \\
Mini Speech Commands & GRU & DeepStellar & 0.87 \\
Mini Speech Commands & LSTM & DeepCover & 0.92 \\
Mini Speech Commands & GRU & DeepCover & 0.93 \\
\bottomrule
\end{tabular*}
\end{table}

\begin{table}[ht]
\caption{The feature importance of extracted features from the SM based on the information gain that they provide on predicting errors of the PM.}\label{tab:feature-importance}
\begin{tabular*}{0.4\textwidth}{@{}ll@{}}
\toprule
Feature & Importance \\ 
\midrule
Final State Share Rate & 0.777352 \\
Count at Final State & 0.133684 \\
New Transitions & 0.050113 \\
Transitions Probability Mean & 0.019410 \\
New States & 0.013005 \\
Trace Probability & 0.006436 \\
\bottomrule
\end{tabular*}
\end{table}

\section{Conclusion}
\label{sec:conclusion}

In this paper, we introduced a method for extracting state machines from RNN-based models to address the following research questions, as mentioned in the Introduction section:
\begin{enumerate}
    \item  How can we effectively extract SMs from RNN-based models while maintaining minimal difference in functionality compared to the PM?
     We proposed an enhanced technique for SM extraction that utilizes k-means clustering on the PM's state vectors to define SM states based on the distribution of states. This provides a more representative view than prior gridding approaches.

    \item How can we evaluate the quality of the extracted SMs?
     We introduced four metrics - Purity, Richness, Goodness and Scale - to quantitatively assess the quality of extracted SMs by comparing to the PM's functionality. To assess the effectiveness of the proposed SM extraction method, we compared it with DeepStellar's approach using the evaluation metrics. Our experimental results show that the proposed state machine extraction algorithm produces high-quality state machines, outperforming the DeepStellar method.

    \item How can we use the explainability achieved through SM extraction to improve the quality of the PM, both in testing and runtime?
     We leveraged the extracted SM to establish six coverage criteria to effectively evaluate test suites for the PM. We utilized the Kolmogorov-Smirnov statistical test to compare the accuracy distributions between areas targeted by the coverage criteria versus overall areas, in order to validate the efficacy of the proposed criteria in measuring test suite quality. We also proposed a tree-based method using SM state features to predict errors in the PM during runtime with over 80\% AUC.

\end{enumerate}

While we have shown promising results on RNN models for image and speech tasks, our approach has some limitations. It is designed for recurrent architectures and has not yet been extended to other neural network models like convolutional or transformer networks. Additionally, the online error prediction depends on having a representative test set with labeled errors to train the decision tree model. Collecting comprehensive error labels poses a practical challenge. Finally, we have focused our evaluation on image classification and speech recognition. Applying these techniques to other intricate sequence modeling tasks, such as translation or text summarization, may surface new issues to address.

As for the future work, an exciting step could be to extend the proposed methodologies to transformer-based models on sequential data. Transformers have shown promising results in various sequential data tasks, such as natural language processing~\cite{vaswani2017attention}. One can broaden the applicability of the suggested techniques to improve the interpretability of transformers, while also employing them to forecast potential inaccuracies in the primary model during runtime.

\section{Code and Data Availability}
\label{sec:data}
In order to facilitate the replication of our experiments and promote open research, we have provided a replication package containing the source code, datasets, and detailed instructions for reproducing our results. The replication package is available on GitHub\footnote{GitHub repository for the replication package: \url{https://github.com/pouriagr/deep-cover}}. We encourage researchers to use and build upon our work, and we welcome any feedback or contributions to improve the methodology and its applications.

% Use the style of numbering in square brackets.
% If nothing is used, default style will be taken.
%\begin{enumerate}[a)]
%\item 
%\item 
%\item 
%\end{enumerate}  

%\begin{itemize}
%\item 
%\item 
%\item 
%\end{itemize}  

%\begin{description}
%\item[]
%\item[] 
%\item[] 
%\end{description}  

% Figure
% \begin{figure}[<options>]
% 	\centering
% 		\includegraphics[<options>]{}
% 	  \caption{}\label{fig1}
% \end{figure}

% \begin{table}[<options>]
% \begin{tabular*}{\tblwidth}{@{}LL@{}}
% \toprule
% \midrule
%  & \\
%  & \\
%  & \\
%  & \\
% \bottomrule
% \end{tabular*}
% \end{table}

% Uncomment and use as the case may be
%\begin{theorem} 
%\end{theorem}

% Uncomment and use as the case may be
%\begin{lemma} 
%\end{lemma}


% \section{}\label{}

% % To print the credit authorship contribution details

% %% Loading bibliography style file
\bibliographystyle{elsarticle-num-names}

% % Loading bibliography database
\bibliography{main}

\begin{thebibliography}{50}
\expandafter\ifx\csname natexlab\endcsname\relax\def\natexlab#1{#1}\fi
\providecommand{\url}[1]{\texttt{#1}}
\providecommand{\href}[2]{#2}
\providecommand{\path}[1]{#1}
\providecommand{\DOIprefix}{doi:}
\providecommand{\ArXivprefix}{arXiv:}
\providecommand{\URLprefix}{URL: }
\providecommand{\Pubmedprefix}{pmid:}
\providecommand{\doi}[1]{\href{http://dx.doi.org/#1}{\path{#1}}}
\providecommand{\Pubmed}[1]{\href{pmid:#1}{\path{#1}}}
\providecommand{\bibinfo}[2]{#2}
\ifx\xfnm\relax \def\xfnm[#1]{\unskip,\space#1}\fi
%Type = Inproceedings
\bibitem[{Graves et~al.(2013)Graves, Mohamed, and Hinton}]{graves2013speech}
\bibinfo{author}{A.~Graves}, \bibinfo{author}{A.-r. Mohamed}, \bibinfo{author}{G.~Hinton},
\newblock \bibinfo{title}{Speech recognition with deep recurrent neural networks},
\newblock in: \bibinfo{booktitle}{2013 IEEE international conference on acoustics, speech and signal processing}, \bibinfo{organization}{Ieee}, \bibinfo{year}{2013}, pp. \bibinfo{pages}{6645--6649}.
%Type = Inproceedings
\bibitem[{Bahdanau et~al.(2015)Bahdanau, Cho, and Bengio}]{bahdanau2014neural}
\bibinfo{author}{D.~Bahdanau}, \bibinfo{author}{K.~Cho}, \bibinfo{author}{Y.~Bengio},
\newblock \bibinfo{title}{Neural machine translation by jointly learning to align and translate},
\newblock in: \bibinfo{booktitle}{Proceedings of the International Conference on Learning Representations (ICLR)}, \bibinfo{year}{2015}.
%Type = Article
\bibitem[{Lipton et~al.(2015)Lipton, Berkowitz, and Elkan}]{lipton2015critical}
\bibinfo{author}{Z.~C. Lipton}, \bibinfo{author}{J.~Berkowitz}, \bibinfo{author}{C.~Elkan},
\newblock \bibinfo{title}{A critical review of recurrent neural networks for sequence learning},
\newblock \bibinfo{journal}{arXiv preprint arXiv:1506.00019}  (\bibinfo{year}{2015}).
%Type = Article
\bibitem[{Huang et~al.(2020)Huang, Kroening, Ruan, Sharp, Sun, Thamo, Wu, and Yi}]{huang2020survey}
\bibinfo{author}{X.~Huang}, \bibinfo{author}{D.~Kroening}, \bibinfo{author}{W.~Ruan}, \bibinfo{author}{J.~Sharp}, \bibinfo{author}{Y.~Sun}, \bibinfo{author}{E.~Thamo}, \bibinfo{author}{M.~Wu}, \bibinfo{author}{X.~Yi},
\newblock \bibinfo{title}{A survey of safety and trustworthiness of deep neural networks: Verification, testing, adversarial attack and defence, and interpretability},
\newblock \bibinfo{journal}{Computer Science Review} \bibinfo{volume}{37} (\bibinfo{year}{2020}) \bibinfo{pages}{100270}.
%Type = Inproceedings
\bibitem[{Du et~al.(2019)Du, Xie, Li, Ma, Liu, and Zhao}]{du2019deepstellar}
\bibinfo{author}{X.~Du}, \bibinfo{author}{X.~Xie}, \bibinfo{author}{Y.~Li}, \bibinfo{author}{L.~Ma}, \bibinfo{author}{Y.~Liu}, \bibinfo{author}{J.~Zhao},
\newblock \bibinfo{title}{Deepstellar: Model-based quantitative analysis of stateful deep learning systems},
\newblock in: \bibinfo{booktitle}{Proceedings of the 2019 27th ACM Joint Meeting on European Software Engineering Conference and Symposium on the Foundations of Software Engineering}, \bibinfo{year}{2019}, pp. \bibinfo{pages}{477--487}.
%Type = Inproceedings
\bibitem[{Chefer et~al.(2021)Chefer, Gur, and Wolf}]{chefer2021transformer}
\bibinfo{author}{H.~Chefer}, \bibinfo{author}{S.~Gur}, \bibinfo{author}{L.~Wolf},
\newblock \bibinfo{title}{Transformer interpretability beyond attention visualization},
\newblock in: \bibinfo{booktitle}{Proceedings of the IEEE/CVF Conference on Computer Vision and Pattern Recognition}, \bibinfo{year}{2021}, pp. \bibinfo{pages}{782--791}.
%Type = Inproceedings
\bibitem[{Barbiero et~al.(2022)Barbiero, Ciravegna, Giannini, Li{\'o}, Gori, and Melacci}]{barbiero2022entropy}
\bibinfo{author}{P.~Barbiero}, \bibinfo{author}{G.~Ciravegna}, \bibinfo{author}{F.~Giannini}, \bibinfo{author}{P.~Li{\'o}}, \bibinfo{author}{M.~Gori}, \bibinfo{author}{S.~Melacci},
\newblock \bibinfo{title}{Entropy-based logic explanations of neural networks},
\newblock in: \bibinfo{booktitle}{Proceedings of the AAAI Conference on Artificial Intelligence}, volume~\bibinfo{volume}{36}, \bibinfo{year}{2022}, pp. \bibinfo{pages}{6046--6054}.
%Type = Inproceedings
\bibitem[{Ayache et~al.(2019)Ayache, Eyraud, and Goudian}]{ayache2019explaining}
\bibinfo{author}{S.~Ayache}, \bibinfo{author}{R.~Eyraud}, \bibinfo{author}{N.~Goudian},
\newblock \bibinfo{title}{Explaining black boxes on sequential data using weighted automata},
\newblock in: \bibinfo{booktitle}{International Conference on Grammatical Inference}, \bibinfo{organization}{PMLR}, \bibinfo{year}{2019}, pp. \bibinfo{pages}{81--103}.
%Type = Article
\bibitem[{Wang et~al.(2022)Wang, Lawrence, and Niepert}]{wang2022state}
\bibinfo{author}{C.~Wang}, \bibinfo{author}{C.~Lawrence}, \bibinfo{author}{M.~Niepert},
\newblock \bibinfo{title}{State-regularized recurrent neural networks to extract automata and explain predictions},
\newblock \bibinfo{journal}{IEEE Transactions on Pattern Analysis and Machine Intelligence}  (\bibinfo{year}{2022}).
%Type = Inproceedings
\bibitem[{Shen et~al.(2018)Shen, Wan, and Chen}]{shen2018munn}
\bibinfo{author}{W.~Shen}, \bibinfo{author}{J.~Wan}, \bibinfo{author}{Z.~Chen},
\newblock \bibinfo{title}{Munn: Mutation analysis of neural networks},
\newblock in: \bibinfo{booktitle}{2018 IEEE International Conference on Software Quality, Reliability and Security Companion (QRS-C)}, \bibinfo{organization}{IEEE}, \bibinfo{year}{2018}, pp. \bibinfo{pages}{108--115}.
%Type = Inproceedings
\bibitem[{Ma et~al.(2018)Ma, Zhang, Sun, Xue, Li, Juefei-Xu, Xie, Li, Liu, Zhao et~al.}]{ma2018deepmutation}
\bibinfo{author}{L.~Ma}, \bibinfo{author}{F.~Zhang}, \bibinfo{author}{J.~Sun}, \bibinfo{author}{M.~Xue}, \bibinfo{author}{B.~Li}, \bibinfo{author}{F.~Juefei-Xu}, \bibinfo{author}{C.~Xie}, \bibinfo{author}{L.~Li}, \bibinfo{author}{Y.~Liu}, \bibinfo{author}{J.~Zhao}, et~al.,
\newblock \bibinfo{title}{Deepmutation: Mutation testing of deep learning systems},
\newblock in: \bibinfo{booktitle}{2018 IEEE 29th International Symposium on Software Reliability Engineering (ISSRE)}, \bibinfo{organization}{IEEE}, \bibinfo{year}{2018}, pp. \bibinfo{pages}{100--111}.
%Type = Article
\bibitem[{Tambon et~al.(2023)Tambon, Khomh, and Antoniol}]{tambon2023probabilistic}
\bibinfo{author}{F.~Tambon}, \bibinfo{author}{F.~Khomh}, \bibinfo{author}{G.~Antoniol},
\newblock \bibinfo{title}{A probabilistic framework for mutation testing in deep neural networks},
\newblock \bibinfo{journal}{Information and Software Technology} \bibinfo{volume}{155} (\bibinfo{year}{2023}) \bibinfo{pages}{107129}.
%Type = Inproceedings
\bibitem[{Pei et~al.(2017)Pei, Cao, Yang, and Jana}]{pei2017deepxplore}
\bibinfo{author}{K.~Pei}, \bibinfo{author}{Y.~Cao}, \bibinfo{author}{J.~Yang}, \bibinfo{author}{S.~Jana},
\newblock \bibinfo{title}{Deepxplore: Automated whitebox testing of deep learning systems},
\newblock in: \bibinfo{booktitle}{proceedings of the 26th Symposium on Operating Systems Principles}, \bibinfo{year}{2017}, pp. \bibinfo{pages}{1--18}.
%Type = Inproceedings
\bibitem[{Tian et~al.(2018)Tian, Pei, Jana, and Ray}]{tian2018deeptest}
\bibinfo{author}{Y.~Tian}, \bibinfo{author}{K.~Pei}, \bibinfo{author}{S.~Jana}, \bibinfo{author}{B.~Ray},
\newblock \bibinfo{title}{Deeptest: Automated testing of deep-neural-network-driven autonomous cars},
\newblock in: \bibinfo{booktitle}{Proceedings of the 40th international conference on software engineering}, \bibinfo{year}{2018}, pp. \bibinfo{pages}{303--314}.
%Type = Inproceedings
\bibitem[{Sun et~al.(2018)Sun, Wu, Ruan, Huang, Kwiatkowska, and Kroening}]{sun2018concolic}
\bibinfo{author}{Y.~Sun}, \bibinfo{author}{M.~Wu}, \bibinfo{author}{W.~Ruan}, \bibinfo{author}{X.~Huang}, \bibinfo{author}{M.~Kwiatkowska}, \bibinfo{author}{D.~Kroening},
\newblock \bibinfo{title}{Concolic testing for deep neural networks},
\newblock in: \bibinfo{booktitle}{Proceedings of the 33rd ACM/IEEE International Conference on Automated Software Engineering}, \bibinfo{year}{2018}, pp. \bibinfo{pages}{109--119}.
%Type = Inproceedings
\bibitem[{Wicker et~al.(2018)Wicker, Huang, and Kwiatkowska}]{wicker2018feature}
\bibinfo{author}{M.~Wicker}, \bibinfo{author}{X.~Huang}, \bibinfo{author}{M.~Kwiatkowska},
\newblock \bibinfo{title}{Feature-guided black-box safety testing of deep neural networks},
\newblock in: \bibinfo{booktitle}{International Conference on Tools and Algorithms for the Construction and Analysis of Systems}, \bibinfo{organization}{Springer}, \bibinfo{year}{2018}, pp. \bibinfo{pages}{408--426}.
%Type = Inproceedings
\bibitem[{Harel-Canada et~al.(2020)Harel-Canada, Wang, Gulzar, Gu, and Kim}]{harel2020neuron}
\bibinfo{author}{F.~Harel-Canada}, \bibinfo{author}{L.~Wang}, \bibinfo{author}{M.~A. Gulzar}, \bibinfo{author}{Q.~Gu}, \bibinfo{author}{M.~Kim},
\newblock \bibinfo{title}{Is neuron coverage a meaningful measure for testing deep neural networks?},
\newblock in: \bibinfo{booktitle}{Proceedings of the 28th ACM Joint Meeting on European Software Engineering Conference and Symposium on the Foundations of Software Engineering}, \bibinfo{year}{2020}, pp. \bibinfo{pages}{851--862}.
%Type = Inproceedings
\bibitem[{Humbatova et~al.(2021)Humbatova, Jahangirova, and Tonella}]{humbatova2021deepcrime}
\bibinfo{author}{N.~Humbatova}, \bibinfo{author}{G.~Jahangirova}, \bibinfo{author}{P.~Tonella},
\newblock \bibinfo{title}{Deepcrime: mutation testing of deep learning systems based on real faults},
\newblock in: \bibinfo{booktitle}{Proceedings of the 30th ACM SIGSOFT International Symposium on Software Testing and Analysis}, \bibinfo{year}{2021}, pp. \bibinfo{pages}{67--78}.
%Type = Article
\bibitem[{Lee and Yannakakis(1996)}]{lee1996principles}
\bibinfo{author}{D.~Lee}, \bibinfo{author}{M.~Yannakakis},
\newblock \bibinfo{title}{Principles and methods of testing finite state machines-a survey},
\newblock \bibinfo{journal}{Proceedings of the IEEE} \bibinfo{volume}{84} (\bibinfo{year}{1996}) \bibinfo{pages}{1090--1123}.
%Type = Article
\bibitem[{Weiss et~al.(2022)Weiss, Goldberg, and Yahav}]{weiss2022extracting}
\bibinfo{author}{G.~Weiss}, \bibinfo{author}{Y.~Goldberg}, \bibinfo{author}{E.~Yahav},
\newblock \bibinfo{title}{Extracting automata from recurrent neural networks using queries and counterexamples (extended version)},
\newblock \bibinfo{journal}{Machine Learning}  (\bibinfo{year}{2022}) \bibinfo{pages}{1--43}.
%Type = Inproceedings
\bibitem[{Okudono et~al.(2020)Okudono, Waga, Sekiyama, and Hasuo}]{okudono2020weighted}
\bibinfo{author}{T.~Okudono}, \bibinfo{author}{M.~Waga}, \bibinfo{author}{T.~Sekiyama}, \bibinfo{author}{I.~Hasuo},
\newblock \bibinfo{title}{Weighted automata extraction from recurrent neural networks via regression on state spaces},
\newblock in: \bibinfo{booktitle}{Proceedings of the AAAI Conference on Artificial Intelligence}, volume~\bibinfo{volume}{34}, \bibinfo{year}{2020}, pp. \bibinfo{pages}{5306--5314}.
%Type = Article
\bibitem[{Weiss et~al.(2018)Weiss, Goldberg, and Yahav}]{weiss2018practical}
\bibinfo{author}{G.~Weiss}, \bibinfo{author}{Y.~Goldberg}, \bibinfo{author}{E.~Yahav},
\newblock \bibinfo{title}{On the practical computational power of finite precision rnns for language recognition},
\newblock \bibinfo{journal}{arXiv preprint arXiv:1805.04908}  (\bibinfo{year}{2018}).
%Type = Article
\bibitem[{Weiss et~al.(2019)Weiss, Goldberg, and Yahav}]{weiss2019learning}
\bibinfo{author}{G.~Weiss}, \bibinfo{author}{Y.~Goldberg}, \bibinfo{author}{E.~Yahav},
\newblock \bibinfo{title}{Learning deterministic weighted automata with queries and counterexamples},
\newblock \bibinfo{journal}{Advances in Neural Information Processing Systems} \bibinfo{volume}{32} (\bibinfo{year}{2019}).
%Type = Inproceedings
\bibitem[{Wei et~al.(2022)Wei, Zhang, and Sun}]{wei2022extracting}
\bibinfo{author}{Z.~Wei}, \bibinfo{author}{X.~Zhang}, \bibinfo{author}{M.~Sun},
\newblock \bibinfo{title}{Extracting weighted finite automata from recurrent neural networks for natural languages},
\newblock in: \bibinfo{booktitle}{Formal Methods and Software Engineering: 23rd International Conference on Formal Engineering Methods, ICFEM 2022, Madrid, Spain, October 24--27, 2022, Proceedings}, \bibinfo{organization}{Springer}, \bibinfo{year}{2022}, pp. \bibinfo{pages}{370--385}.
%Type = Article
\bibitem[{Kolmogorov(1933)}]{kolmogorov1933sulla}
\bibinfo{author}{A.~N. Kolmogorov},
\newblock \bibinfo{title}{Sulla determinazione empirica di una legge di distribuzione},
\newblock \bibinfo{journal}{Giornale dell’Istituto Italiano degli Attuari} \bibinfo{volume}{4} (\bibinfo{year}{1933}) \bibinfo{pages}{83--91}.
%Type = Article
\bibitem[{Angluin(1987)}]{Angluin1987}
\bibinfo{author}{D.~Angluin},
\newblock \bibinfo{title}{Learning regular sets from queries and counterexamples},
\newblock \bibinfo{journal}{Information and Computation} \bibinfo{volume}{75} (\bibinfo{year}{1987}) \bibinfo{pages}{87--106}. \DOIprefix\doi{10.1016/0890-5401(87)90052-6}.
%Type = Article
\bibitem[{Merrill(2019)}]{merrill2019sequential}
\bibinfo{author}{W.~Merrill},
\newblock \bibinfo{title}{Sequential neural networks as automata},
\newblock \bibinfo{journal}{arXiv preprint arXiv:1906.01615}  (\bibinfo{year}{2019}).
%Type = Inproceedings
\bibitem[{Dosovitskiy and Brox(2016)}]{dosovitskiy2016inverting}
\bibinfo{author}{A.~Dosovitskiy}, \bibinfo{author}{T.~Brox},
\newblock \bibinfo{title}{Inverting visual representations with convolutional networks},
\newblock in: \bibinfo{booktitle}{Proceedings of the IEEE conference on computer vision and pattern recognition}, \bibinfo{year}{2016}, pp. \bibinfo{pages}{4829--4837}.
%Type = Inproceedings
\bibitem[{Ma et~al.(2018)Ma, Juefei-Xu, Zhang, Sun, Xue, Li, Chen, Su, Li, Liu et~al.}]{ma2018deepgauge}
\bibinfo{author}{L.~Ma}, \bibinfo{author}{F.~Juefei-Xu}, \bibinfo{author}{F.~Zhang}, \bibinfo{author}{J.~Sun}, \bibinfo{author}{M.~Xue}, \bibinfo{author}{B.~Li}, \bibinfo{author}{C.~Chen}, \bibinfo{author}{T.~Su}, \bibinfo{author}{L.~Li}, \bibinfo{author}{Y.~Liu}, et~al.,
\newblock \bibinfo{title}{Deepgauge: Multi-granularity testing criteria for deep learning systems},
\newblock in: \bibinfo{booktitle}{Proceedings of the 33rd ACM/IEEE International Conference on Automated Software Engineering}, \bibinfo{year}{2018}, pp. \bibinfo{pages}{120--131}.
%Type = Inproceedings
\bibitem[{Kim et~al.(2019)Kim, Feldt, and Yoo}]{kim2019guiding}
\bibinfo{author}{J.~Kim}, \bibinfo{author}{R.~Feldt}, \bibinfo{author}{S.~Yoo},
\newblock \bibinfo{title}{Guiding deep learning system testing using surprise adequacy},
\newblock in: \bibinfo{booktitle}{2019 IEEE/ACM 41st International Conference on Software Engineering (ICSE)}, \bibinfo{organization}{IEEE}, \bibinfo{year}{2019}, pp. \bibinfo{pages}{1039--1049}.
%Type = Inproceedings
\bibitem[{Ross and Doshi-Velez(2018)}]{ross2018improving}
\bibinfo{author}{A.~Ross}, \bibinfo{author}{F.~Doshi-Velez},
\newblock \bibinfo{title}{Improving the adversarial robustness and interpretability of deep neural networks by regularizing their input gradients},
\newblock in: \bibinfo{booktitle}{Proceedings of the AAAI Conference on Artificial Intelligence}, volume~\bibinfo{volume}{32}, \bibinfo{year}{2018}.
%Type = Article
\bibitem[{Rossolini et~al.(2022)Rossolini, Biondi, and Buttazzo}]{rossolini2022increasing}
\bibinfo{author}{G.~Rossolini}, \bibinfo{author}{A.~Biondi}, \bibinfo{author}{G.~Buttazzo},
\newblock \bibinfo{title}{Increasing the confidence of deep neural networks by coverage analysis},
\newblock \bibinfo{journal}{IEEE Transactions on Software Engineering}  (\bibinfo{year}{2022}).
%Type = Article
\bibitem[{Hou et~al.(2022)Hou, Ai, Chen, Yan, Huang, and Chen}]{hou2022similarity}
\bibinfo{author}{R.~Hou}, \bibinfo{author}{S.~Ai}, \bibinfo{author}{Q.~Chen}, \bibinfo{author}{H.~Yan}, \bibinfo{author}{T.~Huang}, \bibinfo{author}{K.~Chen},
\newblock \bibinfo{title}{Similarity-based integrity protection for deep learning systems},
\newblock \bibinfo{journal}{Information Sciences} \bibinfo{volume}{601} (\bibinfo{year}{2022}) \bibinfo{pages}{255--267}.
%Type = Article
\bibitem[{Raghunathan et~al.(2018)Raghunathan, Steinhardt, and Liang}]{raghunathan2018certified}
\bibinfo{author}{A.~Raghunathan}, \bibinfo{author}{J.~Steinhardt}, \bibinfo{author}{P.~Liang},
\newblock \bibinfo{title}{Certified defenses against adversarial examples},
\newblock \bibinfo{journal}{arXiv preprint arXiv:1801.09344}  (\bibinfo{year}{2018}).
%Type = Article
\bibitem[{Antor{\'a}n et~al.(2020)Antor{\'a}n, Allingham, and Hern{\'a}ndez-Lobato}]{antoran2020depth}
\bibinfo{author}{J.~Antor{\'a}n}, \bibinfo{author}{J.~Allingham}, \bibinfo{author}{J.~M. Hern{\'a}ndez-Lobato},
\newblock \bibinfo{title}{Depth uncertainty in neural networks},
\newblock \bibinfo{journal}{Advances in neural information processing systems} \bibinfo{volume}{33} (\bibinfo{year}{2020}) \bibinfo{pages}{10620--10634}.
%Type = Article
\bibitem[{Elman(1990)}]{elman1990finding}
\bibinfo{author}{J.~L. Elman},
\newblock \bibinfo{title}{Finding structure in time},
\newblock \bibinfo{journal}{Cognitive science} \bibinfo{volume}{14} (\bibinfo{year}{1990}) \bibinfo{pages}{179--211}.
%Type = Article
\bibitem[{Hochreiter and Schmidhuber(1997)}]{hochreiter1997long}
\bibinfo{author}{S.~Hochreiter}, \bibinfo{author}{J.~Schmidhuber},
\newblock \bibinfo{title}{Long short-term memory},
\newblock \bibinfo{journal}{Neural computation} \bibinfo{volume}{9} (\bibinfo{year}{1997}) \bibinfo{pages}{1735--1780}.
%Type = Book
\bibitem[{Friedman and Menon(1971)}]{friedman1971fault}
\bibinfo{author}{A.~D. Friedman}, \bibinfo{author}{P.~R. Menon}, \bibinfo{title}{Fault detection in digital circuits}, \bibinfo{publisher}{Prentice Hall}, \bibinfo{year}{1971}.
%Type = Book
\bibitem[{Kohavi and Jha(2009)}]{kohavi2009switching}
\bibinfo{author}{Z.~Kohavi}, \bibinfo{author}{N.~K. Jha}, \bibinfo{title}{Switching and finite automata theory}, \bibinfo{publisher}{Cambridge University Press}, \bibinfo{year}{2009}.
%Type = Inproceedings
\bibitem[{Vassilvitskii and Arthur(2006)}]{vassilvitskii2006k}
\bibinfo{author}{S.~Vassilvitskii}, \bibinfo{author}{D.~Arthur},
\newblock \bibinfo{title}{k-means++: The advantages of careful seeding},
\newblock in: \bibinfo{booktitle}{Proceedings of the eighteenth annual ACM-SIAM symposium on Discrete algorithms}, \bibinfo{year}{2006}, pp. \bibinfo{pages}{1027--1035}.
%Type = Inproceedings
\bibitem[{Cheng et~al.(2018)Cheng, Cao, Xu, and Ma}]{cheng2018manifesting}
\bibinfo{author}{D.~Cheng}, \bibinfo{author}{C.~Cao}, \bibinfo{author}{C.~Xu}, \bibinfo{author}{X.~Ma},
\newblock \bibinfo{title}{Manifesting bugs in machine learning code: An explorative study with mutation testing},
\newblock in: \bibinfo{booktitle}{2018 IEEE International Conference on Software Quality, Reliability and Security (QRS)}, \bibinfo{organization}{IEEE}, \bibinfo{year}{2018}, pp. \bibinfo{pages}{313--324}.
%Type = Article
\bibitem[{Loh(2011)}]{loh2011classification}
\bibinfo{author}{W.-Y. Loh},
\newblock \bibinfo{title}{Classification and regression trees},
\newblock \bibinfo{journal}{Wiley interdisciplinary reviews: data mining and knowledge discovery} \bibinfo{volume}{1} (\bibinfo{year}{2011}) \bibinfo{pages}{14--23}.
%Type = Article
\bibitem[{Quinlan(1986)}]{quinlan1986induction}
\bibinfo{author}{J.~Quinlan},
\newblock \bibinfo{title}{Induction of decision trees. mach. learn}  (\bibinfo{year}{1986}).
%Type = Article
\bibitem[{LeCun et~al.(1998)LeCun, Bottou, Bengio, and Haffner}]{lecun1998gradient}
\bibinfo{author}{Y.~LeCun}, \bibinfo{author}{L.~Bottou}, \bibinfo{author}{Y.~Bengio}, \bibinfo{author}{P.~Haffner},
\newblock \bibinfo{title}{Gradient-based learning applied to document recognition},
\newblock \bibinfo{journal}{Proceedings of the IEEE} \bibinfo{volume}{86} (\bibinfo{year}{1998}) \bibinfo{pages}{2278--2324}.
%Type = Article
\bibitem[{Warden et~al.(2018)Warden, Mueller, Soyer, Alain, and Bengio}]{warden2018speech}
\bibinfo{author}{P.~Warden}, \bibinfo{author}{M.~Mueller}, \bibinfo{author}{H.~Soyer}, \bibinfo{author}{G.~Alain}, \bibinfo{author}{Y.~Bengio},
\newblock \bibinfo{title}{Speech commands: A dataset for limited-vocabulary speech recognition},
\newblock \bibinfo{journal}{arXiv preprint arXiv:1804.03209}  (\bibinfo{year}{2018}).
%Type = Article
\bibitem[{Tieleman et~al.(2012)Tieleman, Hinton et~al.}]{tieleman2012lecture}
\bibinfo{author}{T.~Tieleman}, \bibinfo{author}{G.~Hinton}, et~al.,
\newblock \bibinfo{title}{Lecture 6.5-rmsprop: Divide the gradient by a running average of its recent magnitude},
\newblock \bibinfo{journal}{COURSERA: Neural networks for machine learning} \bibinfo{volume}{4} (\bibinfo{year}{2012}) \bibinfo{pages}{26--31}.
%Type = Misc
\bibitem[{Molnar(2018)}]{molnar2018interprtable}
\bibinfo{author}{C.~Molnar}, \bibinfo{title}{Interprtable machine learning: A guide for making black box models explainable}, \bibinfo{year}{2018}.
%Type = Book
\bibitem[{Jolliffe(2002)}]{jolliffe2002principal}
\bibinfo{author}{I.~T. Jolliffe}, \bibinfo{title}{Principal Component Analysis}, \bibinfo{publisher}{Springer}, \bibinfo{address}{New York, NY, USA}, \bibinfo{year}{2002}.
%Type = Article
\bibitem[{Fisher(1936)}]{fisher1936use}
\bibinfo{author}{R.~A. Fisher},
\newblock \bibinfo{title}{The use of multiple measurements in taxonomic problems},
\newblock \bibinfo{journal}{Annals of Eugenics} \bibinfo{volume}{7} (\bibinfo{year}{1936}) \bibinfo{pages}{179--188}. \DOIprefix\doi{10.1111/j.1469-1809.1936.tb02137.x}.
%Type = Inproceedings
\bibitem[{Vaswani et~al.(2017)Vaswani, Shazeer, Parmar, Uszkoreit, Jones, Gomez, Kaiser, and Polosukhin}]{vaswani2017attention}
\bibinfo{author}{A.~Vaswani}, \bibinfo{author}{N.~Shazeer}, \bibinfo{author}{N.~Parmar}, \bibinfo{author}{J.~Uszkoreit}, \bibinfo{author}{L.~Jones}, \bibinfo{author}{A.~N. Gomez}, \bibinfo{author}{{\L}.~Kaiser}, \bibinfo{author}{I.~Polosukhin},
\newblock \bibinfo{title}{Attention is all you need},
\newblock in: \bibinfo{booktitle}{Advances in Neural Information Processing Systems}, \bibinfo{year}{2017}, pp. \bibinfo{pages}{5998--6008}.

\end{thebibliography}

% % Biography
% % Here goes the biography details.

% % \bio{pic1}
% % Here goes the biography details.

\end{document}